%% file: main.tex
\documentclass[11pt,a4paper]{article}
\usepackage[T1]{fontenc}
\usepackage[american]{babel}
\usepackage[margin=20mm,headheight=14pt]{geometry}
\usepackage{amsmath,amssymb}
\usepackage{newtxtext,newtxmath}
\usepackage{microtype}
\usepackage{graphicx,booktabs,longtable,array,placeins,flafter}
\usepackage{csquotes}
\usepackage[backend=biber,style=apa,natbib=true]{biblatex}
\usepackage{caption}
\DeclareCaptionLabelFormat{arxivfigure}{\textbf{Fig.~#2.}}
\usepackage{fancyhdr}
\usepackage[hidelinks,bookmarksnumbered=true]{hyperref}
\usepackage{bookmark}
\graphicspath{{figures/}}
\AtBeginBibliography{\interlinepenalty=10000}
\fancypagestyle{plain}{\fancyhf{}\fancyfoot[C]{\small\thepage}}
\hypersetup{
 pdftitle={Where Meanings Converge and Diverge: Revisiting Linguistic Relativity Across 81 Languages},
 pdfauthor={Weiran Wang; Wenjuan Qin; Haitao Liu},
 pdfsubject={Linguistic relativity and cross-language conceptual relations},
 pdfkeywords={linguistic relativity, Sapir-Whorf hypothesis, conceptual organization, multilingual comparison, representational similarity}
}
\title{\LARGE\bfseries Where Meanings Converge and Diverge:\\
Revisiting Linguistic Relativity Across 81 Languages}
\author{Weiran Wang, Wenjuan Qin, and Haitao Liu\thanks{Corresponding author: Haitao Liu. Fudan University, Shanghai 200433, China. Email: \href{mailto:htliu@163.com}{htliu@163.com}.}\\[4pt]
\normalsize Fudan University, Shanghai 200433, China}
\date{October 22, 2026}
\begin{document}
\maketitle
\begin{abstract}
Do people who speak different languages organize concepts in similar ways? This question lies at the heart of the Sapir--Whorf hypothesis and debates about linguistic relativity. Shared basic meanings, however, need not imply shared relations among them. We compared relations among 196 basic concepts across 81 languages, using word embeddings from two multilingual models to quantify conceptual structure expressed in language. Cross-language correspondence was greater for sociocultural meanings than for meanings concerning nature and motion, a contrast recurring across models and classifications. Within the sociocultural domain, relational differences increased with geographic separation; less closely related languages differed more across all domains. These findings suggest that universality and relativity are interwoven across conceptual domains. They inform an account in which shared cognitive resources interact with linguistic and local experience, and predict greater cross-language correspondence in human semantic judgments for sociocultural than for nature and motion meanings.
\end{abstract}

\input{introduction}
\input{research_transparency}
\input{methods}
\FloatBarrier
\input{results}
\FloatBarrier
\input{discussion}

\section{Author Contributions}
Conceptualization and methodology: Weiran Wang, Wenjuan Qin, and Haitao Liu. Data curation and formal analysis: Weiran Wang. Visualization and writing---original draft: Weiran Wang, Wenjuan Qin, and Haitao Liu. Writing---review and editing: Wenjuan Qin and Haitao Liu. Supervision: Wenjuan Qin and Haitao Liu.

\section{Acknowledgments}
This paper is a research output of the Digital Humanities and Glottometrics Lab at Fudan University.

\clearpage
\printbibliography[title={References}]
\clearpage
\begin{refsection}
\section*{Supplementary Information}
\addcontentsline{toc}{section}{Supplementary Information}
\noindent\textbf{Where Meanings Converge and Diverge: Revisiting Linguistic Relativity Across 81 Languages}

\medskip
\noindent Weiran Wang, Wenjuan Qin, and Haitao Liu

\noindent Fudan University, Shanghai 200433, China

\medskip
\setcounter{section}{0}
\setcounter{subsection}{0}
\setcounter{table}{0}
\setcounter{figure}{0}
\setcounter{equation}{0}
\setcounter{secnumdepth}{1}
\renewcommand{\thesection}{S\arabic{section}}
\renewcommand{\thetable}{S\arabic{table}}
\renewcommand{\thefigure}{S\arabic{figure}}
\renewcommand{\theequation}{S\arabic{equation}}
\renewcommand{\theHsection}{S\arabic{section}}
\renewcommand{\theHtable}{S\arabic{table}}
\renewcommand{\theHfigure}{S\arabic{figure}}
\renewcommand{\theHequation}{S\arabic{equation}}
\fontsize{10}{12}\selectfont
\renewcommand{\small}{\fontsize{9}{11}\selectfont}
\renewcommand{\scriptsize}{\fontsize{7}{8}\selectfont}
\captionsetup{font=small}
\input{supplementary_information}
\clearpage
\renewcommand{\bibfont}{\small}
\printbibliography[title={Supplementary References}]
\end{refsection}
\end{document}

%% file: introduction.tex
\section{Introduction}

Concepts are mental representations through which people recognize objects, understand events, and generalize beyond what they have directly experienced \parencite{murphy2002concepts,murphy1985role}. Whether people who speak different languages organize these representations in similar ways is a central question about the human mind. Because concepts are expressed through language, linguistic diversity offers a window onto the relationship between shared cognitive capacities and different ways of organizing experience \parencite{evans2009myth,malt2013thought}. The Sapir--Whorf hypothesis, or linguistic relativity, places this relationship at the center of inquiry: how do linguistic distinctions relate to the ways speakers attend to, categorize, and understand the world \parencite{lucy1992language,wolff2011linguistic}? Shared bodily and perceptual capacities provide potential foundations for common conceptual organization \parencite{barsalou2008grounded,spelke2007core}, while language-related differences have been observed in color discrimination, pitch representation, and event categorization \parencite{athanasopoulos2015two,dolscheid_2013,winawer2007russian}. These patterns can coexist: locomotion terms across languages reflect a common biomechanical distinction between walking and running \parencite{malt_2008}, and infants are sensitive to multiple space--pitch mappings before acquiring language-specific conventions \parencite{dolscheid_2014}. Such evidence motivates an account in which shared resources and linguistic experience jointly contribute to meaning organization \parencite{andrews2009integrating,wolff2011linguistic}. This relationship extends beyond the availability of corresponding words. Translation equivalents can refer to corresponding objects or events while participating in different semantic neighborhoods and broader relational structures \parencite{lewis2023local,malt2013thought,thompson2020cultural}. Because relations among meanings contribute to conceptual coherence, categorization, and generalization, their correspondence is itself a psychologically consequential question \parencite{murphy2002concepts,murphy1985role}. A broad assessment of linguistic relativity therefore calls for examining how shared and language-specific relations are distributed across different kinds of meaning.

Word embeddings provide a scalable means of approaching this question by representing regularities in linguistic contexts as relations in a quantitative space \parencite{landauer1997solution,lenci2018distributional}. Their psychological relevance rests on evidence that distributional structure captures aspects of human semantic similarity and perceptual knowledge, alongside complementary information supplied by direct experience \parencite{andrews2009integrating,lewis2019distributional,riordan2011redundancy}. More recently, \textcite{xu2025revealing} showed that representations elicited through concept inference in language-prediction models capture human similarity judgments and category structure, and predict brain responses to object images. These findings support using language-derived representations to investigate aspects of conceptual organization across languages. Previous comparisons have identified shared lexical-network structure and culturally patterned variation \parencite{jackson2019emotion,youn2016universal}. \textcite{thompson2020cultural} found greater semantic alignment in highly structured domains, including number and quantity, than for natural kinds and common actions; \textcite{lewis2023local} further distinguished correspondence within semantic clusters from variability between them. A framework for integrating these findings treats language as a human-driven complex adaptive system, shaped by interacting cognitive, communicative, and environmental constraints \parencite{beckner2009language,liu2014language,liu2018language}. In this framework, linguistic regularities are probabilistic and responsive to patterns of use, making graded and uneven correspondence across meanings theoretically relevant \parencite{beckner2009language,liu2018language}. Here, we compare relations among 196 basic lexical concepts across 81 languages using two multilingual encoders. We ask which contrasts among Biological Core, Nature \& Motion, Socio-Cultural, and Grammar \& Function recur across representational and classificatory choices, and how these patterns relate jointly to geographic distance and estimated language relatedness. The recurring result is greater cross-language correspondence among the sampled Socio-Cultural relations than among Nature \& Motion relations. By locating commonality and variation within a shared basic inventory, this comparison supplies a domain-sensitive pattern for accounts of linguistic relativity to explain. It also predicts a corresponding contrast in matched human semantic-relation judgments, connecting the broad comparison to the psychological organization of concepts.

%% file: research_transparency.tex
\section{Research Transparency Statement}

\subsection{General Disclosures}

\noindent\textit{Conflicts of interest:} The authors declare no competing interests.

\noindent\textit{Funding:} The authors declare no funding.

\noindent\textit{Artificial intelligence:} Multilingual BERT and XLM-R-large were used as research tools to extract lexical representations. Generative AI tools, including GPT-5.6-Luna, Opus-5, and Sonnet-5, assisted with code revision, language editing, drafting support, and LaTeX formatting. All scientific content, methodological decisions, data analyses, and interpretations remain the sole responsibility of the authors, who reviewed and verified all AI-assisted text.

\noindent\textit{Ethics:} This study used publicly available, non-identifiable lexical resources and language metadata. No human participants were recruited, and no new behavioral or neural data were collected.

\subsection{Study Disclosures}

\noindent\textit{Preregistration:} No aspects of this study were preregistered.

\noindent\textit{Materials:} The concept and language inventories, concept mappings, and analysis specifications are provided in the Supplemental Material and accompanying source package, available via Figshare: \url{https://figshare.com/s/63bae7225054e9718549}.

\noindent\textit{Data:} Data and machine-readable summaries supporting the reported analyses of 81 languages and 196 concepts are available via \url{https://figshare.com/s/63bae7225054e9718549}. The accompanying materials document the provenance of the archived model outputs used in the analyses.

\noindent\textit{Analysis scripts:} The accompanying source package provides the analysis scripts and specifications supporting the reported estimates: \url{https://figshare.com/s/63bae7225054e9718549}.

%% file: methods.tex
\section{Method}

\subsection{Languages, Concepts, and Semantic Domains}

We compared relations among translation equivalents in a computational, cross-sectional design. Lexical forms for the starting inventory of 82 languages and 207 basic concepts were downloaded from Wiktionary's language-specific Swadesh lists \parencite{wiktionaryswadeshlists}, following the Swadesh basic-vocabulary tradition \parencite{swadesh1955towards}. Languages were selected for model coverage, usable lexical material, geographic coordinates, and relatedness metadata. This coverage-based sample enables broad comparison without representing a probability sample of the world's languages. No new behavioral or neural responses were collected. Archived lexical forms and Wiktionary concept numbering were retained, including identifiers distinguishing entries with the same English gloss.

Vectors were screened for dimensionality, finite values, zero content, and parseability; the first valid vector was retained for repeated identifiers. Javanese was excluded because identical vectors, rounded to seven decimals, comprised more than half its valid representations (.607). Twelve all-zero observations affected 11 concept identifiers. Excluding these identifiers across languages yielded common support of 81 languages, 196 concepts, and 3,240 unordered language pairs in both encoders. Sample size followed usable common coverage. Complete inventories and exclusions appear in the Supplemental Material.

Four author-defined domains consolidated the topical annotations in the archived study inventory (Table~\ref{tab:domains}). They are operational groupings, rather than independently established conceptual modules. We also used a reference-style grouping that reassigned 22 of the 207 input concepts, including six personal-pronoun items from Biological Core to Grammar \& Function; kinship remained Biological Core. The groupings agreed on 185 input concepts, yielding a complete-case consensus subset of 175 concepts. Item assignments and their sources are documented in the Supplemental Material.

\begin{table}[!htbp]
\centering
\small
\setlength{\tabcolsep}{4pt}
\caption{Operational domains in the primary grouping. Counts refer to the 196 retained concepts; complete item assignments are provided in the Supplemental Material.}
\label{tab:domains}
\begin{tabular}{p{.23\linewidth}rp{.63\linewidth}}
\toprule
Domain & $n$ & Included fields \\
\midrule
Biological Core & 67 & Body, perception, bodily processes, people, kinship, personal pronouns \\
Nature \& Motion & 49 & Nonhuman living kinds, environments, movement, physical change \\
Socio-Cultural & 59 & Properties, evaluation, color, time, actions, material culture, numerals \\
Grammar \& Function & 21 & Demonstratives, interrogatives, quantifiers, relational/function items \\
\bottomrule
\end{tabular}
\end{table}

\subsection{Lexical Representations and Relational Dissimilarity}

We extracted isolated lexical forms with mBERT (\texttt{bert-base-multilingual-cased}) and XLM-R-large (\texttt{xlm-roberta-large}) \parencite{devlin2019bert,conneau2020unsupervised}. Final-layer states were averaged over attended, non-special lexical subwords. Comma-delimited alternatives were encoded separately and averaged equally; forms without usable lexical tokens were missing. Both encoders used the same forms, pooling rules, and support, while retaining their own tokenizers and training histories. Extraction choices matter for interpreting these representations \parencite{vulic2020probing}.

Representational similarity analysis compared the ordering of semantic relations within each language \parencite{kriegeskorte2008representational,beinborn2020semantic}. Vectors were centered on their language-specific mean within each concept set before constructing a cosine-dissimilarity matrix. The full inventory and each domain were centered separately; domain matrices were not subsets of an already centered global matrix. For languages $i$ and $j$ and concept set $C$, cross-language dissimilarity was
\begin{equation}
d_{ij,C}=1-\rho_{\mathrm{Spearman}}\!\left[u(D_{i,C}),u(D_{j,C})\right],
\end{equation}
where $u(D)$ contains upper-triangle entries in matched concept order, excluding the diagonal. Ties received average ranks. Lower values indicate more similarly ordered relations; the index ranges from zero to two and requires no coordinate alignment. Each language pair contributed one global and four domain outcomes per encoder. Concept-pair entries were not independent regression observations. No undefined outcomes occurred. Principal-component projections were descriptive only.

\subsection{Geographic and Linguistic Distance Measures}

Geographic distance was great-circle distance between archived language centroids. Genealogical distance indexed estimated language relatedness using lexical--phonological dissimilarities from the Automated Similarity Judgment Program (ASJP), derived from pointwise mutual-information scores for core-vocabulary forms \parencite{jager2018global}. Higher values indicate more distant estimated relationships. This proxy can reflect inheritance and borrowing and shares material with the semantic inventory; it is not a reconstructed family-tree distance. Both matrices were fixed, symmetric, and complete on the retained languages. Predictors were standardized and entered jointly, so slopes describe associations per standard-deviation increase in one distance, conditional on the other. Geographic separation does not directly measure contact, and word-form dissimilarity does not isolate inheritance.

\subsection{Statistical Analysis and Robustness Checks}

We fitted Gaussian Bayesian regressions with domain-specific intercepts and distance slopes. For each encoder, the domain model was
\begin{equation}
\begin{aligned}
d_{ij,k} &\sim N(\mu_{ij,k},\sigma^2),\\
\mu_{ij,k} &= \alpha_k
 + \beta_{\mathrm{geo},k}z_{\mathrm{geo},ij}
 + \beta_{\mathrm{gen},k}z_{\mathrm{gen},ij}
 + \tfrac12(u_i+u_j).
\end{aligned}
\label{eq:distance-model}
\end{equation}
Here, $i$ and $j$ index languages, $k$ indexes domains, and $z_{\mathrm{geo},ij}$ and $z_{\mathrm{gen},ij}$ are standardized geographic and genealogical distances. The domain-specific intercept $\alpha_k$ describes adjusted dissimilarity; the two $\beta$ terms are total domain slopes. The shared language intercepts $u_i$ and $u_j$ account for repeated participation in language pairs, each with weight one half.

Language effects used a non-centered Gaussian parameterization and a sum-to-zero constraint. Biological Core was the domain reference; other intercepts and slopes were coded as increments. Encoder-specific global and domain fits used 3,240 and 12,960 rows, respectively. A pooled domain model (25,920 rows) included encoder, domain, both predictors, and interactions through Encoder $\times$ Domain $\times$ Predictor, with shared language intercepts and mBERT as the encoder reference. Adjusted means used centered predictors and zero language contribution. Total slopes and contrasts were derived within posterior draws; pooled draws supplied between-encoder comparisons. Models omitted an additional dyad effect and residual correlations across outcomes from the same dyad; intervals are conditional on this specification.

Priors were $N(.9,.2)$ for the reference intercept, $N(0,.2)$ for other fixed coefficients (mean, standard deviation), positive Student-$t(3,0,.5)$ for the language-effect scale, and exponential rate one for residual standard deviation. CmdStan 2.38.0 \parencite{carpenter2017stan} used four No-U-Turn chains, each with 3,000 warm-up and 3,000 retained iterations. Estimates are posterior means with equal-tail 95\% credible intervals (CrIs). Across the complete-support model fits, the maximum $\widehat R$ was 1.008, the minimum bulk effective sample size was 394, and there were no divergent transitions or maximum-tree-depth saturations; the minimum reported E-BFMI was .633. Convergence and posterior predictive checks, full specifications, and coefficients are reported in the Supplemental Material.

We recomputed RSA after regrouping concepts and on the consensus subset, permuted concept labels within languages and domains 2,000 times, and widened intercept/coefficient prior scales from .2 to .5. Permutation probabilities used the add-one correction. Language-omission summaries recomputed means without refitting models; paired-dyad bootstraps did not account for dependence between pairs sharing a language. Secondary English-frequency models used item-level cosine dissimilarity, and archived size/layer checks used earlier vectors; neither constitutes a matched replication of the primary RSA analysis. Full records and extraction provenance are retained in the Supplemental Material.

For illustration, we screened mBERT language triplets for an anchor, a comparison with low genealogical distance but large geographic separation, and a geographically close comparison with higher genealogical distance. We selected the largest positive gap favoring the comparison with lower genealogical distance for Grammar \& Function and the geographically close comparison for Socio-Cultural. These gaps average item-level cosine similarities between language-centered vectors, a different measure from domain RSA. Distance thresholds and complete candidate records are provided in the Supplemental Material.

%% file: results.tex
\section{Results}

\subsection{Semantic Similarity Across Languages Varies by Domain}

\begin{figure}[!htbp]
\centering\includegraphics[width=.98\textwidth]{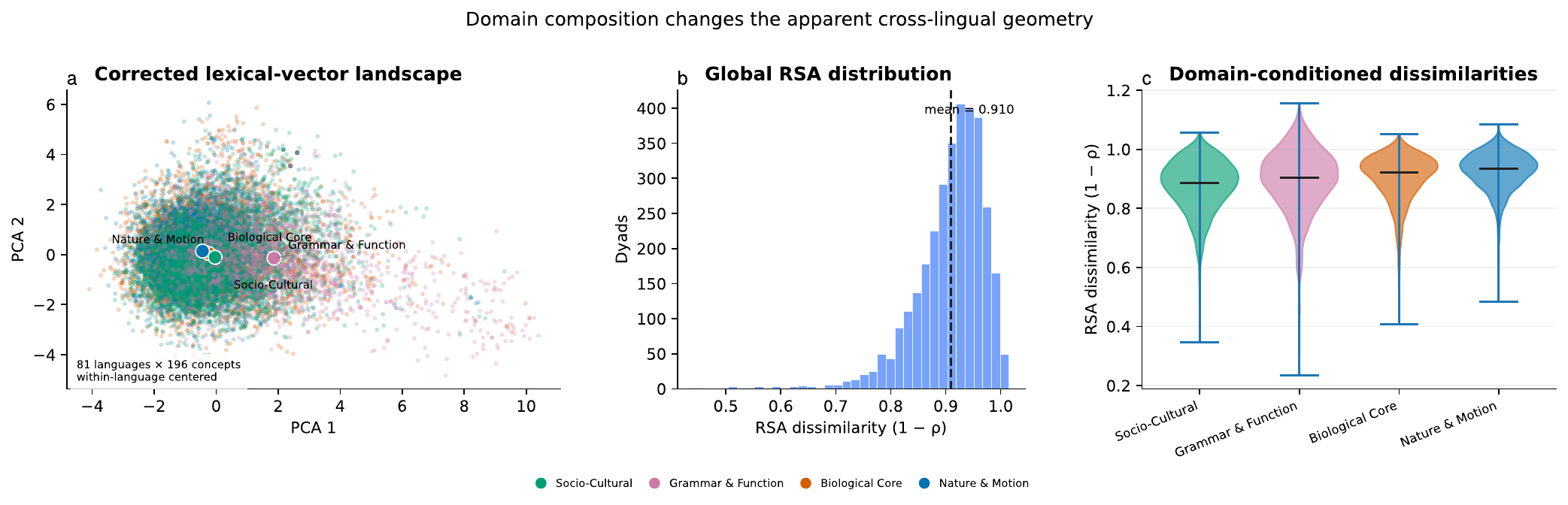}
\caption{\textbf{Cross-language variation in lexical relations.} Multilingual BERT (mBERT) results for 81 languages and 196 concepts. The panels show (a) a descriptive principal-component projection of within-language centered vectors, colored by domain; (b) global representational similarity analysis (RSA) dissimilarities for 3,240 language pairs; and (c) domain distributions. The projection does not establish conceptual modules. Lower dissimilarity indicates more similar ordering of within-language relations; global and domain sets were centered separately.}
\label{fig:global}
\end{figure}

The shared inventory yielded limited overall correspondence but a recurring difference between semantic domains. Across the 3,240 language pairs, adjusted global RSA dissimilarity was .910 (95\% CrI [.909, .911]) in mBERT and .932 [.931, .933] in XLM-R-large. These values describe substantial variation in the ordering of within-language relations, even among translation equivalents. Against this background, we compared the four domains on the same language support (Fig.~\ref{fig:global}).

In mBERT, Socio-Cultural concepts showed the lowest adjusted dissimilarity (.874, 95\% CrI [.872, .877]), whereas Nature \& Motion showed the highest (.924 [.922, .927]). Grammar \& Function (.892 [.890, .895]) and Biological Core (.900 [.898, .903]) occupied intermediate positions. The Nature \& Motion minus Socio-Cultural difference was .0503 [.0470, .0537]. Thus, relations among the sampled Socio-Cultural meanings corresponded more closely across languages than those among Nature \& Motion meanings.

\begin{figure}[!htbp]
\centering\includegraphics[width=.98\textwidth]{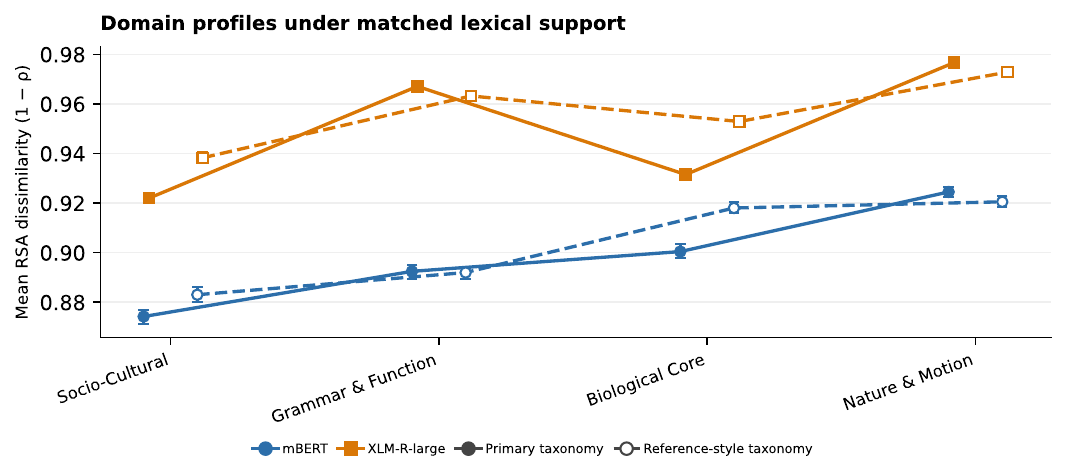}
\caption{\textbf{Domain patterns across encoders and concept groupings.} Points show raw mean representational similarity analysis (RSA) dissimilarities on matched language and concept support. Colors identify encoders; solid lines/filled markers indicate the primary grouping and dashed lines/open markers the reference-style grouping. Error bars show the range of means after omitting each language in turn, not confidence or credible intervals. Adjusted estimates and credible intervals are reported in the text and Supplemental Material.}
\label{fig:cross}
\end{figure}

The same comparison held in XLM-R-large: Socio-Cultural dissimilarity was .922 [.920, .924], compared with .977 [.974, .979] for Nature \& Motion, a difference of .0547 [.0513, .0580]. Biological Core and Grammar \& Function remained intermediate but reversed their order relative to mBERT (Fig.~\ref{fig:cross}; Supplemental Table~\ref{tab:intercepts}). The pooled between-encoder difference in the Nature \& Motion--Socio-Cultural contrast was .0044 [-.0003, .0090]. Agreement therefore concerned the direction of this domain comparison, alongside differences in the complete profile. Even the lower-dissimilarity domains remained far from identical across languages.

Regrouping some concepts preserved this direction. Under the reference-style grouping, raw Socio-Cultural versus Nature \& Motion means were .883 versus .921 in mBERT and .938 versus .973 in XLM-R-large. Restricting the analysis to the 175 concepts assigned identically by both groupings yielded .888 versus .935 and .938 versus .977. Grammar \& Function became the lowest-dissimilarity mBERT group in that subset, qualifying any fixed four-domain ranking. Concept-label calibration also distinguished observed correspondence from random labels: all 16 encoder-by-grouping-by-domain outcomes fell below all 2,000 permutations (Monte Carlo lower-tail $p<.001$). Language omission preserved the descriptive full-inventory orders, and wider priors left the main comparison essentially unchanged; full sensitivity results appear in the Supplemental Material.

\FloatBarrier

\FloatBarrier

\subsection{Semantic Differences Are Associated With Geographic Distance and Language Relatedness}

We examined geographic and genealogical distance jointly using the model in Equation~\ref{eq:distance-model}. Positive slopes indicate greater relational dissimilarity at greater distance, conditional on the other predictor.

\begin{figure}[!ht]
\centering\includegraphics[width=.98\textwidth]{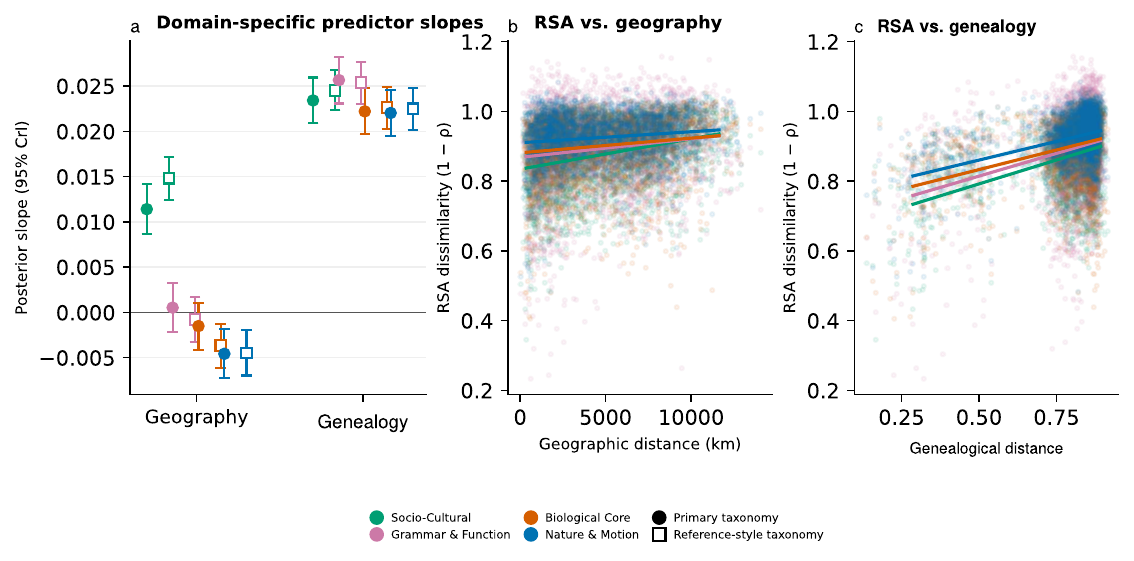}
\caption{\textbf{Geographic and language-relatedness associations.} All panels show multilingual BERT (mBERT). Total domain slopes and 95\% credible intervals are shown in (a); filled circles indicate the primary grouping and open squares the reference-style grouping. Panels (b) and (c) show raw representational similarity analysis (RSA) dyad outcomes and descriptive least-squares lines, distinct from the conditional Bayesian estimates in (a). Genealogical distance is the word-form-based relatedness proxy defined in Method. Complete estimates for both encoders are in Supplemental Tables~\ref{tab:geo_complete} and \ref{tab:slopes}.}
\label{fig:associations}
\end{figure}

Socio-Cultural dissimilarity increased with geographic separation in both encoders: the slope was .0114 (95\% CrI [.0087, .0141]) in mBERT and .0081 [.0055, .0107] in XLM-R-large. Other domains showed less consistent geographic profiles. Nature \& Motion had a negative slope in mBERT (-.0046 [-.0072, -.0019]), while Biological Core had a positive slope in XLM-R-large (.0058 [.0032, .0085]); the remaining intervals included zero (Fig.~\ref{fig:associations}; Supplemental Table~\ref{tab:geo_complete}). Relatively low overall Socio-Cultural dissimilarity therefore coexisted with a recurring geographic gradient. Geography was not associated with a uniform increase across the inventory.

Genealogical distance showed a broader positive association. All four domain slopes were positive in both encoders, and every 95\% CrI lay above zero. Estimates ranged from .02204 to .02566 in mBERT and .01601 to .02107 in XLM-R-large (Supplemental Table~\ref{tab:slopes}). Language pairs with greater estimated genealogical separation also tended to have more dissimilar relational patterns after accounting for geography. This relationship extended across domains, whereas geographic associations varied by domain and encoder. The two predictors thus describe different comparative patterns within the same language pairs.

\FloatBarrier

\subsection{Illustrative Comparisons Among Individual Languages}

\begin{figure}[!htbp]
\centering\includegraphics[width=.98\textwidth,height=.66\textheight,keepaspectratio]{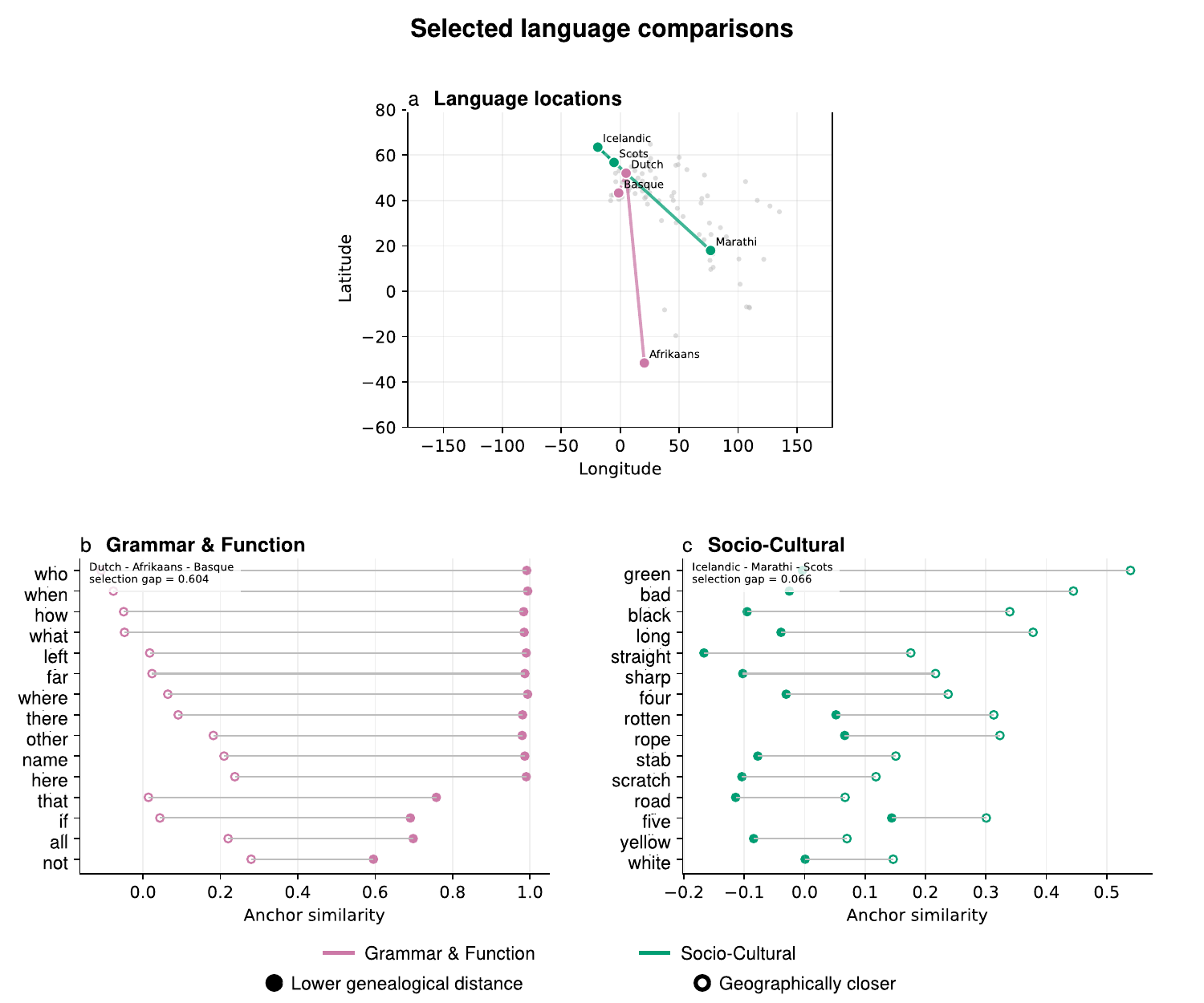}
\caption{\textbf{Selected language comparisons illustrate lexical variation.} Multilingual BERT (mBERT) examples selected for large directional gaps. The locations of the two triplets are shown in (a). Grammar \& Function items in (b) compare Dutch with Afrikaans and Basque; Socio-Cultural items in (c) compare Icelandic with Marathi and Scots. Scores are item-level cosine similarities between language-centered vectors, not domain representational similarity analysis (RSA). Filled markers indicate the comparison with lower genealogical distance and open markers the geographically closer comparison. These selected examples are descriptive.}
\label{fig:triplets}
\end{figure}

Two selected language triplets make these distance configurations concrete at the level of lexical items (Fig.~\ref{fig:triplets}). Each compares an anchor with a geographically distant language having lower genealogical distance and a geographically close language having higher genealogical distance. The examples were selected for large directional gaps in item-level cosine similarity, rather than for their representativeness of all language pairs.

For Grammar \& Function, Dutch was the anchor, Afrikaans the comparison with lower genealogical distance, and Basque the geographically close comparison. Dutch--Afrikaans similarities exceeded Dutch--Basque similarities for many displayed items, including \emph{who}, \emph{when}, \emph{how}, and \emph{what}. In the Socio-Cultural example, Icelandic was compared with Marathi and Scots. Similarities with geographically closer Scots exceeded those with Marathi for selected items including \emph{green}, \emph{long}, \emph{sharp}, \emph{four}, and \emph{five}.

These examples show how spatial separation and word-form relatedness can combine differently in particular comparisons. Their scores concern cross-language similarities of individual lexical vectors, whereas the main analyses compare patterns of within-language relations. They illustrate selected lexical configurations without independently testing the domain-level RSA results or identifying contact and inheritance as causes. Complete selection rules, qualifying candidates, and item records are retained in the Supplemental Material.

\FloatBarrier

%% file: discussion.tex
\section{Discussion}

\subsection{Shared Meanings, Different Relations}

Do people who speak different languages organize basic concepts in similar ways? Our comparison identifies a recurring domain contrast in language-derived representations: relations among the sampled Socio-Cultural meanings corresponded more closely across languages than those among Nature \& Motion meanings. This pattern agrees with \citeauthor{thompson2020cultural}'s (\citeyear{thompson2020cultural}) finding that highly structured meanings can align more closely than natural kinds and common actions. It also complements \citeauthor{lewis2023local}'s (\citeyear{lewis2023local}) distinction between local similarity and broader variability by showing that the extent of correspondence differs across the present groups of basic meanings. The contrast recurred across two encoders and changes in classification and concept support. For linguistic relativity, the contribution is a clearer comparative profile of where relations are relatively shared and where they differ, bringing variation within the lexicon into the language--thought discussion.

This profile connects shared conceptual resources with experience-dependent organization. Grounded-cognition and core-knowledge accounts identify perceptual, bodily, and early representational foundations for understanding the world \parencite{barsalou2008grounded,spelke2007core}. Learning accounts explain how linguistic and experiential information jointly contribute to semantic structure \parencite{andrews2009integrating,riordan2011redundancy}. Our domain contrast fits this combination: common basic meanings can participate in relations whose cross-language correspondence varies with their content. The pitch studies provide a developmental illustration, showing that early sensitivity to multiple mappings can coexist with language-related differences in adults \parencite{dolscheid_2013,dolscheid_2014}. Language-related differences in color discrimination and bilingual event processing likewise show how shared perceptual resources can be deployed differently during particular tasks \parencite{athanasopoulos2015two,winawer2007russian}. Together with evidence that language-prediction models recover human-relevant conceptual structure \parencite{xu2025revealing}, these findings place the present comparison within a psychological account of common resources and learned relations. It extends that account by identifying a recurring contrast across a broad language sample. Conceptual commonality can therefore be examined at more than one level: languages may express corresponding basic meanings while differing in the relations connecting them. This distinction sharpens the comparative question raised by linguistic relativity and links it to theories in which conceptual content depends partly on relations within a larger knowledge system \parencite{malt2013thought,murphy1985role}.

\subsection{Shared Constraints and Local Experience}

The domain contrast is consistent with an interaction between relational systematicity and communicative adaptation. \textcite{thompson2020cultural} associated higher alignment with domains possessing strong internal structure, such as number and quantity. Such structure offers one explanation for correspondence among parts of our Socio-Cultural inventory. Efficient-communication accounts provide a complementary explanation: kinship and color systems balance informative distinctions against representational cost, while color naming also reflects communicative usefulness \parencite{gibson2017color,kemp2012kinship,zaslavsky2018efficient}. Recurrent needs can favor comparable relations among meanings. Locally varying experience can favor different distinctions within the same general framework. Landscape vocabularies vary across languages, ice and snow terminology tracks environmental communicative needs, and semantic variation along shared neurocognitive dimensions is associated with climate \parencite{burenhult2008language,fu2026semantic,regier2016languages}. These findings help explain why relations among Nature \& Motion meanings might vary even when their referents belong to a commonly experienced physical world. Shared perceptual access and shared semantic organization are distinct explanatory issues: environmental regularities provide material for meaning, while experience and use can affect which distinctions and associations become prominent \parencite{barsalou2008grounded,burenhult2008language}. The present contrast thus develops the earlier alignment results into a synthesis in which shared relational constraints and variable experience jointly organize meaning. Greater correspondence in one group and greater variation in another are compatible outcomes of meaning systems adapting to overlapping and differing demands.

The distance associations connect this synthesis to language communities. Socio-Cultural relations were comparatively similar across the sample and also became more dissimilar with geographic separation in both encoders. This combination parallels emotion-semantic research, in which common organizing dimensions coexist with geographically patterned variation \parencite{jackson2019emotion}, and broader alignment findings linking similarity to cultural and geographic proximity \parencite{thompson2020cultural}. Interaction offers a psychologically plausible pathway: selective coordination of task-relevant language is associated with improved joint performance, and language contact can accompany restructuring of color meanings \parencite{fusaroli2012coming,malikmoraleda2023concepts}. Broadly recurring relational patterns can therefore coexist with regional conventions. A domain's average correspondence and its geographic gradient describe different properties: relative commonality across the sample can be accompanied by systematic regional differentiation. This combination specifies how shared and locally patterned organization can coexist within the same field of meanings. Greater relational dissimilarity between more distantly related languages across all domains further situates the profile within historical relationships among languages, extending the historical associations reported by \textcite{thompson2020cultural}. The joint comparison links the organization of meaning to both the kinds of concepts involved and the spatial and historical contexts in which languages are used.

\subsection{Universality and Relativity in an Adaptive System}

The resulting profile gives semantic content to the human-driven complex-adaptive-system perspective introduced above \parencite{beckner2009language,liu2014language,liu2018language}. In this account, cognitive capacities, interaction, and local experience jointly shape linguistic regularities. Cultural-transmission experiments show how learning across generations can produce increasingly structured and learnable systems \parencite{kirby2008cumulative}. Our findings suggest a conceptual patchwork, with relations among meanings woven more similarly across languages in some domains than in others. Shared relational structure and recurrent communicative demands may draw parts of the Socio-Cultural inventory into more similar patterns, whereas variation in ecological experience and communicative needs may allow greater differentiation among Nature \& Motion meanings \parencite{gibson2017color,regier2016languages,thompson2020cultural}. Universality and relativity can thus be understood as interwoven properties of an adaptive meaning system. This probabilistic view connects shared cognitive capacities with locally established patterns of use \parencite{evans2009myth,liu2018language}, offering a domain-sensitive account of how commonality and variation coexist. For the Sapir--Whorf question, this shifts attention toward the distribution of shared and differing relations within conceptual systems. A psychological test follows directly: matched judgments should reveal relatively greater cross-language correspondence for the sampled Socio-Cultural relations than for Nature \& Motion relations. Comparisons with nonverbal materials and manipulations of language use would connect this prediction to the attention, categorization, and learning processes implicated in linguistic relativity \parencite{wolff2011linguistic}.

Several limitations guide that extension. Our evidence comes from model-derived lexical relations; establishing their correspondence to speakers' judgments remains necessary, particularly because naming and nonverbal similarity can dissociate \parencite{malt1999knowing}. Training biases, polysemy, coverage-based sampling, heterogeneous domain assignments, and model-dependent intermediate ranks affect generalization. Communicative, ecological, and historical mechanisms require independent measurements. Within this scope, the study contributes a recurring domain contrast to linguistic relativity and an integrated interpretation linking shared cognitive resources with experience and social use. It makes the relationship between conceptual commonality and linguistic diversity a question about identifiable patterns among meanings, with specific psychological comparisons through which those patterns can be examined.

%% file: supplementary_information.tex
\begingroup
\small
\setlength{\tabcolsep}{3pt}

\section{Sample construction and quality control}

The language inventory was constructed from the full candidate coverage of multilingual BERT available at the time of data assembly. The mBERT model card lists 104 pretraining languages. From this candidate pool, we excluded artificial languages, languages whose available Wikipedia material was predominantly machine-generated, and languages without the geographic or language-history metadata needed for the comparative analysis. Language identifiers were standardized against ISO and Glottolog records, yielding an 82-language study input. Lexical forms were downloaded from the language-specific Swadesh lists indexed by Wiktionary \citep{wiktionaryswadeshlists}. The 207 numeric concept IDs follow Wiktionary's English Swadesh ordering \citep{wiktionaryenglishswadesh}; the archived English concept labels match that sequence after normalization of formatting and parenthetical qualifiers. Thus, 82 languages and 207 concepts describe the study input; this is the inherited input inventory, not a new sampling design introduced by the quality-control procedure. The lexical source is distinct from the Bowern/Concepticon resources subsequently consulted for auxiliary classification. The retrieval dates in the references document source verification and do not establish the original download date. Analyses retained the archived lexical forms rather than replacing them with later online edits.

The archived vector set contained those 207 concept slots in the 82-language input, and the same inventory was extracted from the local mBERT checkpoint on the remote server. Concept identity was reconstructed from numeric concept IDs, not English glosses, because \emph{you} and \emph{man} each occur at two source-list positions. Every vector was screened for parseability, dimensional consistency, finite values, and all-zero content. The quality-control procedure identified 12 all-zero observations (Table~\ref{tab:invalid}). They were treated as invalid rather than transformed by mean centering. Javanese remained degenerate after extraction: its maximum duplicate share at seven-decimal precision was .607, compared with .995 in the archived file. It was excluded under the stated duplicate-share rule. The retained sample therefore contains 81 languages.

For a common support, the primary analysis uses only concepts with valid vectors in all retained languages. Eleven concepts failed this criterion (Table~\ref{tab:excluded}), leaving 196 concepts and 3,240 unordered language dyads. Undefined RSA values remain missing and are never replaced by an arbitrary dissimilarity value.

\section{Representational similarity analysis}

For concept set $C$ in language $L$, vectors were centered within language and set,
\begin{equation}
\hat{v}_{c,L}=v_{c,L}-\frac{1}{|C|}\sum_{i\in C}v_{i,L}.
\end{equation}
Cosine representational distance matrices (RDMs) were constructed from the centered vectors. We call the cross-lingual outcome RSA dissimilarity:
\begin{equation}
d_{\mathrm{sem}}(L_1,L_2)=1-\rho_{\mathrm{Spearman}}\!\left(u(D^{L_1}),u(D^{L_2})\right),
\end{equation}
where $u(\cdot)$ denotes the upper triangle. Because the outcome is $1-\rho$, it can exceed one when the rank correlation is negative. All analyses align concepts by numeric ID and use this RSA definition exclusively.

\section{Primary and reference-style taxonomies}

The primary mapping consolidates the fine-grained topical annotations retained in the project's archived English inventory into four author-defined operational groups. It is the \texttt{combined\_cluster} four-domain classification stored in the source materials. These project annotations and their four-domain consolidation should not be attributed to Wiktionary or treated as the original Bowern list. Its labels are operational summaries rather than homogeneous natural semantic categories. Biological Core includes body, person, kinship, perception, vital processes, and personal pronouns in the primary mapping. Nature \& Motion includes nonhuman living kinds, environments, movement, and physical change. Socio-Cultural contains properties, evaluation, color, time, action and manipulation, material-culture items, and numerals. Grammar \& Function primarily contains demonstratives, interrogatives, quantifiers, and relational or function items. Relevant comparative literature motivates study of these types of meanings \citep{spelke2007core,kemp2012kinship,levinson2003space,burenhult2008language,malt1999knowing,jackson2019emotion,majid2014odors,beckner2009language,christiansen2008language,bybee2010language}, but does not independently validate this particular grouping. The six personal-pronoun items remain Biological Core in the author-defined primary consolidation but move to Grammar \& Function under the reference mapping's explicit functional override. This is an intentional classification difference, not a vector-extraction or analysis bug. Kinship items remain Biological Core under both mappings. Table~\ref{tab:mapping} is the authoritative item-level record.

The alternative mapping is a documented reference-style sensitivity taxonomy constructed using the Bowern-2008-207 list in Concepticon \citep{bowern2008conceptlist,list2016concepticon}, official WOLD and IDS item-level fields \citep{haspelmath2009wold,key2015ids}, WordNet 3.0 lexnames where appropriate \citep{fellbaum1998wordnet}, and explicit functional rules. A study-specific crosswalk aligns the Wiktionary concept slots to Bowern entries through gloss matching and explicit item correspondences; official Concepticon identifiers are then attached through those Bowern entries. This alignment does not make the source inventories identical. Their Concepticon sense labels differ for six aligned items: \emph{child}, \emph{skin}, \emph{smell}, \emph{turn}, \emph{rain}, and \emph{burn}. The frozen Bowern-based crosswalk was retained for this auxiliary sensitivity analysis. The existing \texttt{wold\_field} is retained as a deterministic WOLD/IDS-informed macro-domain rule; the official WOLD semantic field and IDS chapter are stored separately rather than being relabeled as the same variable. The frozen source snapshots supply Bowern-based Concepticon joins for all 207 aligned project items, with linked coverage of 195 Swadesh entries, 188 WOLD entries, and 172 IDS entries. These counts describe the auxiliary crosswalk's coverage; they do not identify the source of the downloaded lexical forms. Missing source coverage is recorded item by item in \url{reanalysis\_20260814/taxonomy\_sources/official\_resource\_crosswalk\_207.tsv}. The two mappings agree on 185 of 207 items. Tables~\ref{tab:agreement}--\ref{tab:mapping} report the mapping comparison, the reassignments, and each concept-level decision.

\section{Literature motivation for the four domains}

\begingroup
\scriptsize
\begin{longtable}{@{}p{.17\textwidth}p{.29\textwidth}p{.36\textwidth}p{.08\textwidth}@{}}
\caption{Literature motivation and support for the four operational domains. Counts refer first to the 207-item input and then to the 196-concept complete-case support.}\label{tab:domain_literature}\\
\toprule
Domain & Semantic focus & Representative comparative and cognitive literature & Input / complete \\
\midrule
\endfirsthead
\toprule
Domain & Semantic focus & Representative comparative and cognitive literature & Input / complete \\
\midrule
\endhead
\midrule
\multicolumn{4}{r}{Continued on next page} \\
\endfoot
\bottomrule
\endlastfoot
Biological Core & Body, person, kinship, perception, vital processes, and primary-mapping personal pronouns & Core knowledge, embodied structure, kinship systems, and recurrent lexical organization \citep{spelke2007core,kemp2012kinship,pagel2009human} & 68 / 67 \\
Nature \& Motion & Nonhuman living kinds, physical environments, movement, and physical change & Frames of reference, motion descriptions, landscape language, and spatial cognition \citep{levinson2003space,burenhult2008language,talmy1985lexicalization,levinson2006grammars} & 53 / 49 \\
Socio-Cultural & Properties, evaluation, colour, time, actions, material culture, and numerals & Artifact naming, cultural categorization, emotion, odor, and communicative coordination \citep{malt1999knowing,malt2013thought,jackson2019emotion,majid2014odors,thompson2020cultural} & 60 / 59 \\
Grammar \& Function & Demonstratives, interrogatives, relations, quantifiers, and other function items & Grammaticalization, discourse functions, usage-based structure, and language--cognition interaction \citep{beckner2009language,christiansen2008language,croft2001radical,bybee2010language,levinson2006grammars} & 26 / 21 \\
\end{longtable}
\endgroup

The counts in the final column are descriptive support counts, not evidence for the conceptual validity of the domains. The RSA-blind source/ontology audit found 89.4\% exact agreement across the 207 input concepts, Cohen's $\kappa=.856$ (2,000-concept bootstrap 95\% interval [.794, .908]), and macro-F1 $=.885$ ([.835, .927]). The endpoint contrast was then tested under the reference mapping and consensus subset; those outcome-level checks support the broad Nature \& Motion versus Socio-Cultural direction but do not establish a unique four-domain taxonomy. The literature supplies the motivation for the grouping; the complete-case rule then determines which items enter the comparable RSA support.

\section{Geographic distance and genealogical relatedness}

The geographic and genealogical-relatedness matrices were frozen as quality-controlled input files for the analysis. Geographic values are great-circle distances between archived language centroids. We operationalized genealogical relatedness with an ASJP lexical--phonological distance derived from core-vocabulary PMI scores \citep{church1990word,jager2018global}. Throughout this article, genealogical distance refers to this ASJP-based operational measure. It serves as an approximate relatedness measure rather than a formal phylogenetic-tree distance. Because ASJP and the semantic analysis both derive from core-vocabulary traditions, they share partial material overlap. The input hashes in Table~\ref{tab:provenance} verify the exact matrices used by the remote runs.

\section{Comparative models}\label{sec:comparative-models}

The retained multiple-membership model is
\begin{equation}
d_{\mathrm{sem},ij}=\alpha_k+\beta_{\mathrm{geo},k}z_{\mathrm{geo},ij}+\beta_{\mathrm{gen},k}z_{\mathrm{gen},ij}+\tfrac{1}{2}(u_i+u_j)+\epsilon_{ij}.
\end{equation}
Equal language-membership weights sum to one. The Gaussian language effects were represented with a non-centered parameterization and an explicit sum-to-zero constraint, which identifies the intercept separately from the average repeated language contribution. The centered intercept prior was $Normal(.9,.2)$, coefficient priors were $Normal(0,.2)$, the language-effect standard deviation had a positive Student-t distribution with degrees of freedom 3, location 0, and scale .5 prior, and residual $\sigma$ had an $Exponential(1)$ prior. Four independent chains used 3,000 warm-up and 3,000 retained iterations with seed base 42, $\texttt{adapt\_delta}=.99$, and $\texttt{max\_treedepth}=12$. Sampling was performed on the remote server with the complete 3,240-dyad support; no data subsampling was used. Since the four domain outcomes for one dyad are correlated, the model intervals are conditional on this residual specification; the dyad-cluster bootstrap below preserves within-dyad domain dependence but does not eliminate dependence between different dyads sharing a language.

Convergence is evaluated on substantive model parameters after excluding Stan sampler-diagnostic columns such as $\texttt{treedepth\_\_}$; the full and filtered values are retained in the machine-readable diagnostics. The model-specific R-hat, effective sample size, divergence, treedepth, E-BFMI, and posterior-predictive results are reported in Table~\ref{tab:diagnostics}. Tables~\ref{tab:global}--\ref{tab:slopes} provide complete summaries.

The pooled model concatenates the mBERT and XLM-R-large domain rows and adds a model indicator, Model $\times$ Domain, Model $\times$ Geography, Model $\times$ ASJP, and corresponding three-way interactions. It retains the same repeated-language structure and priors. Its model-interaction posteriors, rather than a visual comparison of separate credible intervals, are the formal test of cross-model differences; the complete coefficient table is archived with the machine-readable output.

The Bayesian specification uses four chains with 3,000 warm-up and 3,000 retained iterations per chain, $\texttt{adapt\_delta}=.99$, seed base 42, and $\texttt{max\_treedepth}=12$. All model-by-model and pooled manuscript fits use the complete data support and these settings; no reduced-iteration fit enters a reported estimate. The settings and acceptance diagnostics are recorded in the provenance table and machine-readable manifests.

\begin{longtable}{@{}p{.38\textwidth}p{.12\textwidth}p{.20\textwidth}p{.22\textwidth}@{}}
\caption{Formal pooled cross-model posterior contrasts. The mBERT Biological Core intercept is the reference. Intervals are equal-tail 95\% credible intervals from the full-support pooled model.}\label{tab:pooled}\\
\toprule
Term & Mean & 95\% CrI & Interpretation \\
\midrule
\endfirsthead
\toprule
Term & Mean & 95\% CrI & Interpretation \\
\midrule
\endhead
\midrule
\multicolumn{4}{r}{Continued on next page} \\
\endfoot
\bottomrule
\endlastfoot
Primary; Model (XLM-R-large) & $0.0311$ & $[0.0277,0.0345]$ & Model main effect relative to mBERT Biological Core. \\
Primary; Model $\times$ Nature \& Motion & $0.0211$ & $[0.0162,0.0260]$ & Model-by-domain increment. \\
Primary; Model $\times$ Socio-Cultural & $0.0167$ & $[0.0119,0.0213]$ & Model-by-domain increment. \\
Primary; Model $\times$ Grammar \& Function & $0.0435$ & $[0.0387,0.0483]$ & Model-by-domain increment. \\
Primary; Model $\times$ Geography & $0.0060$ & $[0.0023,0.0096]$ & Difference in baseline geography slopes. \\
Primary; Model $\times$ ASJP & $-0.0052$ & $[-0.0088,-0.0015]$ & Difference in baseline ASJP slopes. \\
Primary; Model $\times$ Nature \& Motion $\times$ Geography & $-0.0051$ & $[-0.0104,0.0000]$ & Domain-specific geography-slope difference. \\
Primary; Model $\times$ Socio-Cultural $\times$ Geography & $-0.0107$ & $[-0.0158,-0.0055]$ & Domain-specific geography-slope difference. \\
Primary; Model $\times$ Grammar \& Function $\times$ Geography & $-0.0100$ & $[-0.0152,-0.0048]$ & Domain-specific geography-slope difference. \\
Primary; Model $\times$ Nature \& Motion $\times$ ASJP & $-0.0030$ & $[-0.0082,0.0021]$ & Domain-specific ASJP-slope difference. \\
Primary; Model $\times$ Socio-Cultural $\times$ ASJP & $-0.0036$ & $[-0.0087,0.0015]$ & Domain-specific ASJP-slope difference. \\
Primary; Model $\times$ Grammar \& Function $\times$ ASJP & $-0.0085$ & $[-0.0136,-0.0033]$ & Domain-specific ASJP-slope difference. \\
Primary; Derived endpoint contrast & $0.0044$ & $[-0.0003,0.0090]$ & XLM-R-large minus mBERT endpoint contrast. \\
External; Model (XLM-R-large) & $0.0349$ & $[0.0319,0.0379]$ & Model main effect relative to mBERT Biological Core. \\
External; Model $\times$ Nature \& Motion & $0.0175$ & $[0.0132,0.0218]$ & Model-by-domain increment. \\
External; Model $\times$ Socio-Cultural & $0.0204$ & $[0.0161,0.0246]$ & Model-by-domain increment. \\
External; Model $\times$ Grammar \& Function & $0.0364$ & $[0.0321,0.0407]$ & Model-by-domain increment. \\
External; Model $\times$ Geography & $0.0017$ & $[-0.0015,0.0049]$ & Difference in baseline geography slopes. \\
External; Model $\times$ ASJP & $-0.0075$ & $[-0.0107,-0.0043]$ & Difference in baseline ASJP slopes. \\
External; Model $\times$ Nature \& Motion $\times$ Geography & $-0.0009$ & $[-0.0054,0.0037]$ & Domain-specific geography-slope difference. \\
External; Model $\times$ Socio-Cultural $\times$ Geography & $-0.0106$ & $[-0.0152,-0.0061]$ & Domain-specific geography-slope difference. \\
External; Model $\times$ Grammar \& Function $\times$ Geography & $-0.0031$ & $[-0.0077,0.0015]$ & Domain-specific geography-slope difference. \\
External; Model $\times$ Nature \& Motion $\times$ ASJP & $-0.0075$ & $[-0.0108,-0.0042]$ & Domain-specific ASJP-slope difference. \\
External; Model $\times$ Socio-Cultural $\times$ ASJP & $-0.0111$ & $[-0.0144,-0.0079]$ & Domain-specific ASJP-slope difference. \\
External; Model $\times$ Grammar \& Function $\times$ ASJP & $-0.0088$ & $[-0.0120,-0.0056]$ & Domain-specific ASJP-slope difference. \\
External; Derived endpoint contrast & $-0.0029$ & $[-0.0071,0.0014]$ & XLM-R-large minus mBERT endpoint contrast. \\
\end{longtable}

The prior-scale sensitivity used the same complete support and full four-chain specification with coefficient prior scale $0.5$ instead of $0.2$. The derived endpoint contrast was $0.0044$ [$-0.0005,0.0091$] under the primary taxonomy and $-0.0029$ [$-0.0071,0.0014$] under the reference-style taxonomy. These estimates are numerically indistinguishable from the default-scale results for the inferential claim: the cross-model endpoint difference is not taxonomy-invariant.

\paragraph{Paired language-dyad sensitivity.}
The same unordered language dyad contributes observations to both encoders, so we also resampled complete dyads as blocks, retaining all four domains and both models together. This sensitivity analysis preserves the paired encoder and domain observations within each dyad, but does not eliminate dependence between different dyads sharing a language. It does not replace the full-support pooled posterior. The XLM-R-large minus mBERT difference in the Nature~\& Motion minus Socio-Cultural endpoint contrast was $0.0044$ (95\% paired-bootstrap interval $[0.0012,0.0074]$) under the primary taxonomy, but $-0.0029$ (95\% interval $[-0.0062,0.0002]$) under the reference-style taxonomy. Thus, within-model endpoint contrasts can be stable while the cross-model difference of those contrasts remains taxonomy-sensitive. The complete four-domain paired summaries and manifests are archived in \texttt{reanalysis\_20260814/paired\_cross\_model\_bootstrap\_20260815/} and \texttt{reanalysis\_20260814/paired\_cross\_model\_bootstrap\_external\_20260815/}; each uses all 3,240 dyads and 2,000 replicates.

\begin{table}[ht]
\centering
\small
\caption{Paired language-dyad bootstrap for the cross-model endpoint contrast. The estimand is $(\text{XLM-R-large}-\text{mBERT})$ applied to the Nature~\& Motion minus Socio-Cultural contrast. Both models and all four domains are resampled together within each of 2,000 complete-dyad replicates.}
\label{tab:pairedcrossmodel}
\begin{tabular}{lrrr}
\toprule
Taxonomy & Full estimate & 2.5\% & 97.5\% \\
\midrule
Primary & 0.0044 & 0.0012 & 0.0074 \\
Reference-style taxonomy & $-0.0029$ & $-0.0062$ & 0.0002 \\
\bottomrule
\end{tabular}
\end{table}

\section{Cross-model and representation sensitivity}

Within mBERT, both full observed inventories yield the ordering Socio-Cultural $<$ Grammar \& Function $<$ Biological Core $<$ Nature \& Motion. A 2,000-resample dyad-cluster bootstrap preserves that ordering in 100.0\% of primary-taxonomy resamples and 99.75\% of reference-style taxonomy resamples (2,000 and 1,995 resamples, respectively). The positive Socio-Cultural-by-geography interaction excludes zero under both mappings. The mBERT Grammar-by-ASJP interaction is .00346 [.000015, .00698] under the primary mapping and .00277 [-.00036, .00591] under the alternative mapping. XLM-R-large does not reproduce a consistently positive grammatical increment. Accordingly, the evidence supports a broad positive ASJP association, not a uniquely grammatical-functional association.

XLM-R-large extraction used the same lexical items and processing decisions (Table~\ref{tab:cross-extraction}). It preserves the observed-inventory Socio-Cultural endpoint as lowest and Nature \& Motion as highest under both taxonomies, but Biological Core and Grammar \& Function exchange rank relative to mBERT (Tables~\ref{tab:cross-domain}--\ref{tab:cross-order}). The full four-domain ordering is therefore checkpoint-dependent. Dyad-level agreement is moderate overall and strongest for Socio-Cultural concepts (Table~\ref{tab:cross-agreement}). All leave-one-language-out runs preserve each model's observed-inventory ordering, so the middle-domain swap is not driven by a single language.

As a separate coverage check, we evaluated the XLM-R RSA using the screened language-QC file as the inclusion rule rather than the 82-name source inventory. The run retained 81 languages, 196 complete-case concepts, all 3,240 dyads, and 32,400 rows across two taxonomies and five outcomes (global plus four domains), with no sampling. Table~\ref{tab:fullvalid} exactly reproduces the fixed-sample XLM-R output. This documents coverage and provenance rather than providing an independent-data replication; the same support, vectors, and matrices are used.

\begin{table}[ht]
\centering
\scriptsize
\setlength{\tabcolsep}{2pt}
\caption{Full-valid XLM-R coverage check. Entries are raw RSA dissimilarity means for the same 81-language, 196-concept, 3,240-dyad support; no Bayesian intervals are assigned to this coverage description.}\label{tab:fullvalid}
\begin{tabular}{lrrrrr}
\toprule
Taxonomy & All & Biological & Nature & Socio-Cultural & Grammar \\
\midrule
Primary & .9321 & .9315 & .9767 & .9220 & .9671 \\
External & .9321 & .9529 & .9729 & .9383 & .9632 \\
\bottomrule
\end{tabular}
\end{table}

The layer-wise RSA analysis was computed on the same 81-language, 196-concept complete-case support for all 13 archived mBERT layers (Table~\ref{tab:layers}). It is reported as a representation-level sensitivity analysis, not as a second confirmatory model. The frequency-adjusted item analysis was likewise rebuilt from the extracted vectors and numeric concept IDs. Its English Zipf values come from the archived SUBTLEX/\texttt{wordfreq}-based table and are used as a broad item-level control rather than as language-specific corpus frequencies (Table~\ref{tab:frequency}). The mBERT model uses all 635,040 concept--dyad observations (196 concepts $\times$ 3,240 language dyads), analytically marginalized multiple-membership language and concept random intercepts, and exact Gaussian restricted maximum likelihood. Optimization converged with a full-rank 13-column fixed-effect design; no dyad sampling was used.

\section{Concept-label permutation calibration}

For each encoder, taxonomy, and domain, we also permuted concept labels within the language-level vector arrays 2,000 times while retaining the complete 81-language support. The observed mean RSA dissimilarity was lower than every permutation replicate for all 16 encoder--taxonomy--domain combinations. With the $+1$ Monte Carlo correction, the correctly directed lower-tail probability was $p_{\mathrm{lower}}=.0005$ and the two-sided probability was $p_{\mathrm{two-sided}}=.0010$; the upper-tail probability was 1.0. This calibration tests whether the observed concept correspondence is compatible with random label alignment. It does not test the causal interpretation of the geographic or ASJP associations, nor does it make the dyad-level RSA entries independent.

\section{Consensus-subset sensitivity}

The source and ontology assignments agree for 185 of the 207 concepts. The consensus RSA uses the same 81-language support and vectors. Because the extraction has 11 invalid concept IDs, the complete-case RSA support contains 175 concepts (53 Biological Core, 47 Nature \& Motion, 55 Socio-Cultural, and 20 Grammar \& Function). The two-domain contrast remains (without implying that Socio-Cultural remains the minimum in both encoders): Socio-Cultural is lower than Nature \& Motion for both mBERT (.888 versus .935) and XLM-R-large (.938 versus .977). The complete middle-domain ordering is not invariant: mBERT gives Grammar \& Function $<$ Socio-Cultural $<$ Biological Core $<$ Nature \& Motion, whereas XLM-R-large gives Socio-Cultural $<$ Biological Core $<$ Grammar \& Function $<$ Nature \& Motion. Thus the consensus subset supports the Socio-Cultural--Nature \& Motion contrast but qualifies any claim about the full four-domain hierarchy. The machine-readable consensus RSA outputs and concept-coverage records are included in \texttt{reanalysis\_20260814/}.

\section{Dependence and domain-size sensitivity}\label{sec:domain-size}

The following archived mBERT sensitivity checks use 81 languages and 196 concepts. Their manifest identifies the earlier archived vector directory, rather than the fresh vectors used by the final Bayesian analysis. They are retained for provenance transparency and are not presented as matched fresh-vector, dual-encoder validation. We resampled the 3,240 unordered dyads with replacement 2,000 times, retaining all four domain outcomes within each sampled dyad. This cluster bootstrap produces percentile intervals for domain means, pairwise differences, and rank stability. We then repeatedly sampled without replacement from every domain to equalize its concept count: 21 concepts per domain under the primary mapping and 30 under the reference-style mapping. We recomputed RSA for 500 samples per taxonomy. In the equal-size analysis, the Socio-Cultural--Nature \& Motion ordering was retained in 89.6\% of primary and 95.8\% of reference-style taxonomy samples, whereas the full four-domain ordering occurred in only 33.6\% and 35.4\%, respectively. These results qualify the interpretation of the observed-inventory ranking: the endpoint contrast is more stable than the complete ordering. Machine-readable outputs and the exact scripts are in \texttt{reanalysis\_20260814/}; the primary analyses use the complete support, and the resampling procedures are explicitly labeled sensitivity analyses.

\section{Detailed robustness results}
\label{sec:robustness}

This section provides the complete numerical results for the robustness checks summarized in the main text Results section. All checks address the stability of the domain profile and the cross-model comparison under different analytical choices.

\subsection{Reference-style classification}

The reference-style classification reassigns 22 of the 207 input concepts. Under this alternative mapping, the adjusted Socio-Cultural versus Nature \& Motion means were .883 versus .921 for mBERT and .938 versus .973 for XLM-R-large. The within-model Nature \& Motion minus Socio-Cultural contrasts were .0375 (95\% CrI [.0345, .0405]) for mBERT and .0346 [.0316, .0376] for XLM-R-large. The cross-model difference in these contrasts was $-.0029$ ([-.0071, .0014]). The endpoint contrast is therefore preserved, but the reassignment changes the separation among domains and the order of the middle positions. Complete domain means and credible intervals under the reference-style classification are reported in SI Table~\ref{tab:intercepts}.

\subsection{Leave-one-language-out summaries}

Across all 81 leave-one-language-out summaries, each multilingual language model retained its full-inventory domain ordering under both the primary and reference-style classifications. The minimum and maximum means obtained by omitting each language in turn are displayed in the main text as error bars in the cross-model domain profile figure. No single language drove the observed Socio-Cultural--Nature \& Motion contrast. The complete leave-one-out range summaries are provided in SI Table~\ref{tab:cross-domain}.

\subsection{Paired dyad bootstrap}

The paired language-dyad bootstrap resampled complete dyads with replacement, retaining both encoders and all four domains. This preserves within-dyad pairing while leaving dependence among dyads sharing a language. Its cross-model Nature \& Motion--Socio-Cultural difference was .0044 (95\% interval [.0012, .0074]) under the primary taxonomy and $-.0029$ ([-.0062, .0002]) under the reference-style taxonomy. The full-support pooled posterior intervals include zero under both taxonomies (.0044 [-.0003, .0090] and $-.0029$ [-.0071, .0014], respectively). Changing the coefficient-prior scale from .2 to .5 did not alter those pooled conclusions. The bootstrap and pooled posterior describe different uncertainty specifications; the bootstrap cannot establish an invariant cross-model difference. Paired-bootstrap archive locations are specified in Section~\ref{sec:comparative-models}.

\subsection{Frequency-adjusted secondary models}

The frequency-adjusted analyses use all 635,040 concept--dyad observations for each encoder. They use item-level cosine dissimilarity, rather than domain-level RSA, and an English Zipf covariate, rather than language-specific corpus frequencies. The frequency coefficient was negative in both fits (mBERT $-.02620$, 95\% Wald interval [-.03228, -.02012]; XLM-R-large $-.01414$, [-.02424, -.00404]). Several domain main-effect intervals include zero. Because the outcome and estimands differ from the primary RSA analysis, these models do not show that the RSA domain contrast is unchanged after frequency adjustment or independent of word frequency. Specifications and coefficients are in Table~\ref{tab:frequency}.

\subsection{Layer diagnostics and equal-size subsampling}

Archived equal-size subsampling results are reported in Section~\ref{sec:domain-size}. Table~\ref{tab:layers} reports Grammar \& Function and Socio-Cultural outcomes across archived mBERT layers. These diagnostics describe layer dependence in those two domains on the archived-vector support. The table does not report Nature \& Motion and therefore does not establish that the Socio-Cultural--Nature \& Motion contrast persists through layers 6--12 or is strongest in the final layer. As specified in Section~\ref{sec:domain-size}, the archived checks are not matched fresh-vector, dual-encoder validation.

\subsection{Concept-level profiles and triplet diagnostics}

Concept-level profiles and selected language triplets serve as descriptive illustrations rather than mechanism tests. Triplet diagnostics identified language triples that passed geographic and operational genealogical distance filters while exhibiting the stated directional domain contrast. The complete diagnostic inventory is archived in \texttt{reanalysis\_20260814/secondary\_fulltriplets\_20260815/}. These diagnostics illustrate particular cases but do not constitute independent statistical tests of the domain profile.

\subsection{Summary}

The robustness checks collectively support the Socio-Cultural--Nature \& Motion endpoint contrast and the positive ASJP association across both models. The complete four-domain ordering and some geographic associations, however, remain sensitive to domain assignment, the multilingual language model, and the specific concepts included in each domain. The endpoint contrast is comparatively stable, whereas the exact ordering of all four domains is more conditional on analytical choices.

\section{Triplet diagnostics and complete candidate inventory}

The main-text triplet figure shows two selected mBERT configurations under two stated filters: genealogically-close/geographically-far versus geographically-close/genealogically-far, with the corresponding directional domain contrast. To provide the complete diagnostic record, the secondary-analysis script enumerated every unique language triplet passing those filters and the directional contrast, without a top-case cap. The machine-readable inventory is archived under \url{reanalysis\_20260814/secondary\_fulltriplets\_20260815/}: \texttt{triplet\_all\_filter\_candidates.csv} records every distance-filtered candidate, \texttt{triplet\_all\_qualifying.csv} records every candidate with the stated directional contrast, and \texttt{triplet\_all\_qualifying\_word\_details.csv} gives the concept-level support gaps for every qualifying triplet. The archived top-five files are retained as secondary-analysis inputs.

\begin{table}[ht]
\centering
\scriptsize
\setlength{\tabcolsep}{2pt}
\caption{Dependence- and domain-size-sensitivity results. Percentages are proportions of resamples or equal-size concept subsamples. The dyad bootstrap retains all four domain outcomes within each sampled dyad.}\label{tab:sensitivity}
\resizebox{\columnwidth}{!}{%
\begin{tabular}{@{}llrrrr@{}}
\toprule
Analysis & Taxonomy & Full (\%) & S$<$N (\%) & S min. (\%) & N max. (\%) \\
\midrule
Dyad bootstrap & Primary & 100.0 & 100.0 & 100.0 & 100.0 \\
Dyad bootstrap & Reference-style & 99.75 & 100.0 & 100.0 & 99.75 \\
Equal-size concepts & Primary & 33.6 & 89.6 & 65.4 & 70.8 \\
Equal-size concepts & Reference-style & 35.4 & 95.8 & 69.4 & 54.4 \\
\bottomrule
\end{tabular}
}
\end{table}

\section{Reproducibility materials}

The reproducibility materials include the concept mapping, observation- and language-level quality-control records, the complete-case concept list, archived mBERT vectors, mBERT extraction metadata and quality-control records, RSA summaries, sensitivity analyses, XLM-R-large extraction metadata and RSA summaries, fixed-effect summaries, diagnostics, posterior-predictive summaries, source code, seeds, and hashes. Checkpoint weights are not included in the manuscript package. Available tokenizer hashes, software versions, extraction rules, and extraction commands are recorded. The XLM-R-large extraction manifest marks its weight file as not hashed; complete weight-hash coverage is therefore not claimed.
\section{Supplementary Tables}

\begin{longtable}{rlllp{.25\textwidth}}
\caption{Invalid concept--language observations identified before RSA. All 12 were all-zero vectors in the extraction and were treated as missing.}\label{tab:invalid}\\
\toprule
ID & Concept & Language & Status & Domain \\
\midrule
\endfirsthead
\toprule
ID & Concept & Language & Status & Domain \\
\midrule
\endhead
\midrule
\multicolumn{5}{r}{Continued on next page} \\
\endfoot
\bottomrule
\endlastfoot
201 & at & Japanese & all-zero-vector & Grammar \& Function \\
202 & in & Japanese & all-zero-vector & Grammar \& Function \\
203 & with & Japanese & all-zero-vector & Grammar \& Function \\
204 & and & Japanese & all-zero-vector & Grammar \& Function \\
206 & because & Japanese & all-zero-vector & Grammar \& Function \\
58 & bark & Marathi & all-zero-vector & Nature \& Motion \\
58 & bark & Nepali & all-zero-vector & Nature \& Motion \\
125 & stand & Occitan & all-zero-vector & Nature \& Motion \\
192 & dull & Occitan & all-zero-vector & Socio-Cultural \\
148 & moon & Thai & all-zero-vector & Nature \& Motion \\
149 & star & Thai & all-zero-vector & Nature \& Motion \\
2 & you & Vietnamese & all-zero-vector & Biological Core \\
\end{longtable}

\begin{longtable}{rllrp{.32\textwidth}}
\caption{Concepts excluded by the constant-coverage rule. A concept was retained only when valid in all 81 nondegenerate languages.}\label{tab:excluded}\\
\toprule
ID & Concept & Primary domain & Valid & Invalid observations \\
\midrule
\endfirsthead
\toprule
ID & Concept & Primary domain & Valid & Invalid observations \\
\midrule
\endhead
\midrule
\multicolumn{5}{r}{Continued on next page} \\
\endfoot
\bottomrule
\endlastfoot
2 & you & Biological Core & 80 & Vietnamese:all-zero-vector \\
58 & bark & Nature \& Motion & 79 & Marathi:all-zero-vector; Nepali:all-zero-vector \\
125 & stand & Nature \& Motion & 80 & Occitan:all-zero-vector \\
148 & moon & Nature \& Motion & 80 & Thai:all-zero-vector \\
149 & star & Nature \& Motion & 80 & Thai:all-zero-vector \\
192 & dull & Socio-Cultural & 80 & Occitan:all-zero-vector \\
201 & at & Grammar \& Function & 80 & Japanese:all-zero-vector \\
202 & in & Grammar \& Function & 80 & Japanese:all-zero-vector \\
203 & with & Grammar \& Function & 80 & Japanese:all-zero-vector \\
204 & and & Grammar \& Function & 80 & Japanese:all-zero-vector \\
206 & because & Grammar \& Function & 80 & Japanese:all-zero-vector \\
\end{longtable}

\begin{longtable}{lrrrr}
\caption{Agreement between the primary and reference-style concept taxonomies. Bootstrap intervals use 2,000 concept-level resamples.}\label{tab:agreement}\\
\toprule
Statistic & Estimate & 2.5\% & 97.5\% & $B$ \\
\midrule
\endfirsthead
\toprule
Statistic & Estimate & 2.5\% & 97.5\% & $B$ \\
\midrule
\endhead
\midrule
\multicolumn{5}{r}{Continued on next page} \\
\endfoot
\bottomrule
\endlastfoot
agreement & 0.894 & 0.850 & 0.932 & 2000 \\
cohen\_kappa & 0.856 & 0.794 & 0.908 & 2000 \\
macro\_f1 & 0.885 & 0.835 & 0.927 & 2000 \\
\end{longtable}

\begin{longtable}{lrr}
\caption{Concept counts by taxonomy and domain before the complete-case exclusion of 11 concepts.}\label{tab:counts}\\
\toprule
Domain & Primary & External \\
\midrule
\endfirsthead
\toprule
Domain & Primary & External \\
\midrule
\endhead
\midrule
\multicolumn{3}{r}{Continued on next page} \\
\endfoot
\bottomrule
\endlastfoot
Biological Core & 68 & 53 \\
Nature \& Motion & 53 & 54 \\
Socio-Cultural & 60 & 64 \\
Grammar \& Function & 26 & 36 \\
\end{longtable}

\begin{longtable}{r p{.12\textwidth} p{.16\textwidth} p{.16\textwidth} p{.26\textwidth} l}
\caption{The 22 concepts assigned to different domains by the reference-style mapping.}\label{tab:reassign}\\
\toprule
ID & Concept & Primary & External & Basis & Confidence \\
\midrule
\endfirsthead
\toprule
ID & Concept & Primary & External & Basis & Confidence \\
\midrule
\endhead
\midrule
\multicolumn{6}{r}{Continued on next page} \\
\endfoot
\bottomrule
\endlastfoot
1 & I & Biological Core & Grammar \& Function & functional\_override & high \\
2 & you & Biological Core & Grammar \& Function & functional\_override & high \\
3 & he,she,it & Biological Core & Grammar \& Function & functional\_override & medium \\
4 & we & Biological Core & Grammar \& Function & functional\_override & high \\
5 & you & Biological Core & Grammar \& Function & functional\_override & medium \\
6 & they & Biological Core & Grammar \& Function & functional\_override & high \\
22 & one & Socio-Cultural & Grammar \& Function & functional\_override & high \\
23 & two & Socio-Cultural & Grammar \& Function & functional\_override & high \\
24 & three & Socio-Cultural & Grammar \& Function & functional\_override & high \\
25 & four & Socio-Cultural & Grammar \& Function & functional\_override & high \\
103 & know & Biological Core & Socio-Cultural & WordNet/Bowern source alignment & medium \\
104 & think & Biological Core & Socio-Cultural & WordNet/Bowern source alignment & medium \\
106 & fear & Biological Core & Socio-Cultural & WordNet/Bowern source alignment & medium \\
112 & hunt & Nature \& Motion & Socio-Cultural & WordNet/Bowern source alignment & medium \\
139 & count & Biological Core & Grammar \& Function & functional\_override & medium \\
140 & say & Biological Core & Socio-Cultural & WordNet/Bowern source alignment & medium \\
141 & sing & Biological Core & Socio-Cultural & WordNet/Bowern source alignment & medium \\
142 & play & Nature \& Motion & Socio-Cultural & WordNet/Bowern source alignment & medium \\
145 & freeze & Biological Core & Nature \& Motion & WordNet/Bowern source alignment & high \\
146 & swell & Biological Core & Nature \& Motion & WordNet/Bowern source alignment & high \\
169 & burn & Biological Core & Nature \& Motion & WordNet/Bowern source alignment & medium \\
207 & name & Grammar \& Function & Socio-Cultural & WordNet/Bowern source alignment & medium \\
\end{longtable}

\begin{longtable}{lrrr}
\caption{Dyad-level agreement of RSA dissimilarities between taxonomies.}\label{tab:dyadagreement}\\
\toprule
Domain & Dyads & Spearman $\rho$ & Mean absolute difference \\
\midrule
\endfirsthead
\toprule
Domain & Dyads & Spearman $\rho$ & Mean absolute difference \\
\midrule
\endhead
\midrule
\multicolumn{4}{r}{Continued on next page} \\
\endfoot
\bottomrule
\endlastfoot
Biological Core & 3240 & 0.928 & 0.024 \\
Nature \& Motion & 3240 & 0.956 & 0.014 \\
Socio-Cultural & 3240 & 0.934 & 0.025 \\
Grammar \& Function & 3240 & 0.764 & 0.051 \\
\end{longtable}

\begin{longtable}{lrr}
\caption{Global RSA multiple-membership models. Global intercepts summarize all concepts and are not Biological Core domain intercepts.}\label{tab:global}\\
\toprule
Encoder & Mean & 95\% CrI\\\midrule\endfirsthead
\toprule
Encoder & Mean & 95\% CrI\\\midrule\endhead
\bottomrule\endfoot
mbert & 0.90973 & [0.90867, 0.91079] \\
xlmr & 0.93211 & [0.93085, 0.93338] \\
\end{longtable}

\begin{longtable}{lllp{.34\textwidth}}
\caption{Complete fixed effects for both encoders. Domain indices 1, 2, and 3 denote Nature \& Motion, Socio-Cultural, and Grammar \& Function; Biological Core is the reference.}\label{tab:fixed}\\
\toprule
Classification & Encoder & Parameter & Estimate [95\% CrI]\\\midrule\endfirsthead
\toprule
Classification & Encoder & Parameter & Estimate [95\% CrI]\\\midrule\endhead
\bottomrule\endfoot
Primary & mBERT & alpha & 0.90043 [0.89813, 0.90274] \\
Primary & mBERT & beta\_geo & -0.00155 [-0.00426, 0.00113] \\
Primary & mBERT & beta\_asjp & 0.02220 [0.01963, 0.02478] \\
Primary & mBERT & beta\_domain[1] & 0.02405 [0.02076, 0.02728] \\
Primary & mBERT & beta\_domain[2] & -0.02623 [-0.02951, -0.02292] \\
Primary & mBERT & beta\_domain[3] & -0.00798 [-0.01118, -0.00476] \\
Primary & mBERT & beta\_geo\_domain[1] & -0.00305 [-0.00656, 0.00044] \\
Primary & mBERT & beta\_geo\_domain[2] & 0.01293 [0.00940, 0.01643] \\
Primary & mBERT & beta\_geo\_domain[3] & 0.00206 [-0.00143, 0.00556] \\
Primary & mBERT & beta\_asjp\_domain[1] & -0.00016 [-0.00366, 0.00335] \\
Primary & mBERT & beta\_asjp\_domain[2] & 0.00122 [-0.00232, 0.00475] \\
Primary & mBERT & beta\_asjp\_domain[3] & 0.00346 [0.00001, 0.00698] \\
Primary & XLM-R-large & alpha & 0.93150 [0.92926, 0.93373] \\
Primary & XLM-R-large & beta\_geo & 0.00583 [0.00319, 0.00848] \\
Primary & XLM-R-large & beta\_asjp & 0.02107 [0.01855, 0.02357] \\
Primary & XLM-R-large & beta\_domain[1] & 0.04520 [0.04198, 0.04838] \\
Primary & XLM-R-large & beta\_domain[2] & -0.00952 [-0.01274, -0.00627] \\
Primary & XLM-R-large & beta\_domain[3] & 0.03560 [0.03248, 0.03881] \\
Primary & XLM-R-large & beta\_geo\_domain[1] & -0.00820 [-0.01170, -0.00479] \\
Primary & XLM-R-large & beta\_geo\_domain[2] & 0.00226 [-0.00113, 0.00562] \\
Primary & XLM-R-large & beta\_geo\_domain[3] & -0.00803 [-0.01148, -0.00457] \\
Primary & XLM-R-large & beta\_asjp\_domain[1] & -0.00319 [-0.00666, 0.00020] \\
Primary & XLM-R-large & beta\_asjp\_domain[2] & -0.00242 [-0.00586, 0.00103] \\
Primary & XLM-R-large & beta\_asjp\_domain[3] & -0.00506 [-0.00855, -0.00164] \\
Alternative & mBERT & alpha & 0.91802 [0.91590, 0.92010] \\
Alternative & mBERT & beta\_geo & -0.00365 [-0.00604, -0.00117] \\
Alternative & mBERT & beta\_asjp & 0.02264 [0.02033, 0.02493] \\
Alternative & mBERT & beta\_domain[1] & 0.00249 [-0.00039, 0.00538] \\
Alternative & mBERT & beta\_domain[2] & -0.03500 [-0.03797, -0.03205] \\
Alternative & mBERT & beta\_domain[3] & -0.02605 [-0.02895, -0.02314] \\
Alternative & mBERT & beta\_geo\_domain[1] & -0.00083 [-0.00392, 0.00227] \\
Alternative & mBERT & beta\_geo\_domain[2] & 0.01844 [0.01532, 0.02159] \\
Alternative & mBERT & beta\_geo\_domain[3] & 0.00283 [-0.00033, 0.00595] \\
Alternative & mBERT & beta\_asjp\_domain[1] & -0.00015 [-0.00326, 0.00298] \\
Alternative & mBERT & beta\_asjp\_domain[2] & 0.00189 [-0.00125, 0.00502] \\
Alternative & mBERT & beta\_asjp\_domain[3] & 0.00277 [-0.00036, 0.00591] \\
Alternative & XLM-R-large & alpha & 0.95292 [0.95093, 0.95485] \\
Alternative & XLM-R-large & beta\_geo & 0.00085 [-0.00142, 0.00312] \\
Alternative & XLM-R-large & beta\_asjp & 0.01932 [0.01715, 0.02149] \\
Alternative & XLM-R-large & beta\_domain[1] & 0.01998 [0.01721, 0.02273] \\
Alternative & XLM-R-large & beta\_domain[2] & -0.01464 [-0.01739, -0.01188] \\
Alternative & XLM-R-large & beta\_domain[3] & 0.01031 [0.00757, 0.01313] \\
Alternative & XLM-R-large & beta\_geo\_domain[1] & -0.00166 [-0.00464, 0.00128] \\
Alternative & XLM-R-large & beta\_geo\_domain[2] & 0.00781 [0.00486, 0.01079] \\
Alternative & XLM-R-large & beta\_geo\_domain[3] & -0.00031 [-0.00329, 0.00267] \\
Alternative & XLM-R-large & beta\_asjp\_domain[1] & -0.00013 [-0.00310, 0.00280] \\
Alternative & XLM-R-large & beta\_asjp\_domain[2] & -0.00178 [-0.00473, 0.00118] \\
Alternative & XLM-R-large & beta\_asjp\_domain[3] & 0.00141 [-0.00154, 0.00441] \\
\end{longtable}

\begin{longtable}{lllp{.34\textwidth}}
\caption{Adjusted mean: both encoders and both classifications. Posterior means and equal-tail 95\% credible intervals reconstructed draw by draw from the frozen full-support fits.}\label{tab:intercepts}\\
\toprule
Classification & Encoder & Domain & Estimate [95\% CrI]\\\midrule\endfirsthead
\toprule
Classification & Encoder & Domain & Estimate [95\% CrI]\\\midrule\endhead
\bottomrule\endfoot
Primary & mBERT & Biological Core & 0.90043 [0.89813, 0.90274] \\
Primary & mBERT & Nature \& Motion & 0.92448 [0.92215, 0.92683] \\
Primary & mBERT & Socio-Cultural & 0.87420 [0.87186, 0.87655] \\
Primary & mBERT & Grammar \& Function & 0.89245 [0.89015, 0.89473] \\
Primary & XLM-R-large & Biological Core & 0.93150 [0.92926, 0.93373] \\
Primary & XLM-R-large & Nature \& Motion & 0.97669 [0.97447, 0.97895] \\
Primary & XLM-R-large & Socio-Cultural & 0.92197 [0.91970, 0.92427] \\
Primary & XLM-R-large & Grammar \& Function & 0.96710 [0.96487, 0.96934] \\
Alternative & mBERT & Biological Core & 0.91802 [0.91590, 0.92010] \\
Alternative & mBERT & Nature \& Motion & 0.92050 [0.91844, 0.92256] \\
Alternative & mBERT & Socio-Cultural & 0.88301 [0.88093, 0.88506] \\
Alternative & mBERT & Grammar \& Function & 0.89197 [0.88993, 0.89399] \\
Alternative & XLM-R-large & Biological Core & 0.95292 [0.95093, 0.95485] \\
Alternative & XLM-R-large & Nature \& Motion & 0.97290 [0.97093, 0.97487] \\
Alternative & XLM-R-large & Socio-Cultural & 0.93828 [0.93630, 0.94021] \\
Alternative & XLM-R-large & Grammar \& Function & 0.96323 [0.96128, 0.96516] \\
\end{longtable}

\begin{longtable}{lllp{.34\textwidth}}
\caption{Geography: both encoders and both classifications. Posterior means and equal-tail 95\% credible intervals reconstructed draw by draw from the frozen full-support fits.}\label{tab:geo_complete}\\
\toprule
Classification & Encoder & Domain & Estimate [95\% CrI]\\\midrule\endfirsthead
\toprule
Classification & Encoder & Domain & Estimate [95\% CrI]\\\midrule\endhead
\bottomrule\endfoot
Primary & mBERT & Biological Core & -0.00155 [-0.00426, 0.00113] \\
Primary & mBERT & Nature \& Motion & -0.00459 [-0.00723, -0.00187] \\
Primary & mBERT & Socio-Cultural & 0.01138 [0.00865, 0.01409] \\
Primary & mBERT & Grammar \& Function & 0.00051 [-0.00225, 0.00321] \\
Primary & XLM-R-large & Biological Core & 0.00583 [0.00319, 0.00848] \\
Primary & XLM-R-large & Nature \& Motion & -0.00236 [-0.00505, 0.00031] \\
Primary & XLM-R-large & Socio-Cultural & 0.00810 [0.00546, 0.01072] \\
Primary & XLM-R-large & Grammar \& Function & -0.00219 [-0.00485, 0.00049] \\
Alternative & mBERT & Biological Core & -0.00365 [-0.00604, -0.00117] \\
Alternative & mBERT & Nature \& Motion & -0.00448 [-0.00692, -0.00211] \\
Alternative & mBERT & Socio-Cultural & 0.01479 [0.01236, 0.01722] \\
Alternative & mBERT & Grammar \& Function & -0.00082 [-0.00322, 0.00158] \\
Alternative & XLM-R-large & Biological Core & 0.00085 [-0.00142, 0.00312] \\
Alternative & XLM-R-large & Nature \& Motion & -0.00082 [-0.00307, 0.00140] \\
Alternative & XLM-R-large & Socio-Cultural & 0.00865 [0.00634, 0.01092] \\
Alternative & XLM-R-large & Grammar \& Function & 0.00053 [-0.00174, 0.00280] \\
\end{longtable}
\begin{longtable}{lllp{.34\textwidth}}
\caption{Genealogical distance: both encoders and both classifications. Posterior means and equal-tail 95\% credible intervals reconstructed draw by draw from the frozen full-support fits.}\label{tab:slopes}\\
\toprule
Classification & Encoder & Domain & Estimate [95\% CrI]\\\midrule\endfirsthead
\toprule
Classification & Encoder & Domain & Estimate [95\% CrI]\\\midrule\endhead
\bottomrule\endfoot
Primary & mBERT & Biological Core & 0.02220 [0.01963, 0.02478] \\
Primary & mBERT & Nature \& Motion & 0.02204 [0.01947, 0.02463] \\
Primary & mBERT & Socio-Cultural & 0.02342 [0.02085, 0.02603] \\
Primary & mBERT & Grammar \& Function & 0.02566 [0.02315, 0.02820] \\
Primary & XLM-R-large & Biological Core & 0.02107 [0.01855, 0.02357] \\
Primary & XLM-R-large & Nature \& Motion & 0.01788 [0.01540, 0.02038] \\
Primary & XLM-R-large & Socio-Cultural & 0.01865 [0.01616, 0.02120] \\
Primary & XLM-R-large & Grammar \& Function & 0.01601 [0.01339, 0.01849] \\
Alternative & mBERT & Biological Core & 0.02264 [0.02033, 0.02493] \\
Alternative & mBERT & Nature \& Motion & 0.02250 [0.02020, 0.02476] \\
Alternative & mBERT & Socio-Cultural & 0.02453 [0.02220, 0.02683] \\
Alternative & mBERT & Grammar \& Function & 0.02541 [0.02310, 0.02769] \\
Alternative & XLM-R-large & Biological Core & 0.01932 [0.01715, 0.02149] \\
Alternative & XLM-R-large & Nature \& Motion & 0.01919 [0.01702, 0.02134] \\
Alternative & XLM-R-large & Socio-Cultural & 0.01755 [0.01540, 0.01974] \\
Alternative & XLM-R-large & Grammar \& Function & 0.02074 [0.01858, 0.02292] \\
\end{longtable}

\begin{longtable}{lrrrrrr}
\caption{Sampling diagnostics for the non-centered, sum-to-zero multiple-membership models. R-hat and ESS summarize substantive model parameters; sampler-diagnostic columns are retained separately in the machine-readable output. The pooled rows use the complete 25,920-row support.}\label{tab:diagnostics}\\
\toprule
Model & $N$ & Max $\widehat R$ & Min ESS & Div. & Depth & Min E-BFMI \\
\midrule
\endfirsthead
\toprule
Model & $N$ & Max $\widehat R$ & Min ESS & Div. & Depth & Min E-BFMI \\
\midrule
\endhead
\midrule
\multicolumn{7}{r}{Continued on next page} \\
\endfoot
\bottomrule
\endlastfoot
mBERT global (primary) & 3240 & 1.006 & 434 & 0 & 0 & 0.642 \\
XLM-R global (primary) & 3240 & 1.003 & 644 & 0 & 0 & 0.633 \\
mBERT domain (primary) & 12960 & 1.006 & 533 & 0 & 0 & 0.704 \\
XLM-R domain (primary) & 12960 & 1.005 & 951 & 0 & 0 & 0.648 \\
mBERT global (external) & 3240 & 1.006 & 434 & 0 & 0 & 0.642 \\
XLM-R global (external) & 3240 & 1.003 & 644 & 0 & 0 & 0.633 \\
mBERT domain (external) & 12960 & 1.008 & 394 & 0 & 0 & 0.689 \\
XLM-R domain (external) & 12960 & 1.004 & 1172 & 0 & 0 & 0.689 \\
Pooled (primary; scale .2) & 25920 & 1.005 & 745 & 0 & 0 & 0.700 \\
Pooled (primary; scale .5) & 25920 & 1.004 & 486 & 0 & 0 & 0.695 \\
Pooled (external; scale .2) & 25920 & 1.006 & 559 & 0 & 0 & 0.718 \\
Pooled (external; scale .5) & 25920 & 1.008 & 573 & 0 & 0 & 0.697 \\
\end{longtable}

All eight model-by-model rows and all four pooled rows are completed four-chain, 3,000-warm-up/3,000-retained model fits on the complete support. Sampling diagnostics showed zero divergent transitions and zero maximum-depth saturations; the table reports the minimum E-BFMI across chains and the minimum bulk ESS among substantive parameters. For the reference-style XLM-R domain analysis, the adjusted posterior means were .953 [.951, .955] for Biological Core, .973 [.971, .975] for Nature \& Motion, .938 [.936, .940] for Socio-Cultural, and .963 [.961, .965] for Grammar \& Function. These formal intervals support the endpoint contrast but do not make the middle-domain order encoder-invariant. The pooled endpoint contrast is $0.0044$ [$-0.0003,0.0090$] under the primary taxonomy and $-0.0029$ [$-0.0071,0.0014$] under the reference-style taxonomy; the prior-scale sensitivity leaves both conclusions unchanged. The fixed-sample coverage check remains descriptive.

\begingroup
\scriptsize
\setlength{\tabcolsep}{1.5pt}
\begin{longtable}{rlrrrrr}
\caption{Layer-wise RSA sensitivity analysis for Grammar \& Function and Socio-Cultural. Every row uses the same 81 languages, 196 complete-case concepts, and 3,240 dyads. Slopes and 95\% ordinary least-squares intervals are descriptive and are not additional confirmatory tests.}\label{tab:layers}\\
\toprule
Layer & Domain & Mean & Geo. & 95\% CI & ASJP & 95\% CI \\
\midrule
\endfirsthead
\toprule
Layer & Domain & Mean & Geo. & 95\% CI & ASJP & 95\% CI \\
\midrule
\endhead
\midrule
\multicolumn{7}{r}{Continued on next page} \\
\endfoot
\bottomrule
\endlastfoot
0 & Grammar \& Function & 0.8789 & -0.0002 & [-0.0042, 0.0038] & 0.0285 & [0.0245, 0.0324] \\
0 & Socio-Cultural & 0.8646 & 0.0133 & [0.0099, 0.0168] & 0.0303 & [0.0268, 0.0338] \\
1 & Grammar \& Function & 0.8891 & -0.0049 & [-0.0087, -0.0011] & 0.0273 & [0.0235, 0.0311] \\
1 & Socio-Cultural & 0.8701 & 0.0121 & [0.0088, 0.0154] & 0.0314 & [0.0281, 0.0347] \\
2 & Grammar \& Function & 0.8876 & -0.0048 & [-0.0087, -0.0009] & 0.0294 & [0.0255, 0.0332] \\
2 & Socio-Cultural & 0.8608 & 0.0135 & [0.0101, 0.0169] & 0.0339 & [0.0305, 0.0373] \\
3 & Grammar \& Function & 0.8749 & -0.0049 & [-0.0088, -0.0009] & 0.0327 & [0.0288, 0.0367] \\
3 & Socio-Cultural & 0.8464 & 0.0158 & [0.0122, 0.0193] & 0.0362 & [0.0326, 0.0397] \\
4 & Grammar \& Function & 0.8648 & -0.0020 & [-0.0061, 0.0021] & 0.0332 & [0.0291, 0.0373] \\
4 & Socio-Cultural & 0.8351 & 0.0169 & [0.0132, 0.0206] & 0.0369 & [0.0332, 0.0406] \\
5 & Grammar \& Function & 0.8662 & 0.0004 & [-0.0037, 0.0046] & 0.0325 & [0.0284, 0.0366] \\
5 & Socio-Cultural & 0.8289 & 0.0168 & [0.0130, 0.0205] & 0.0364 & [0.0327, 0.0402] \\
6 & Grammar \& Function & 0.8688 & 0.0007 & [-0.0033, 0.0047] & 0.0315 & [0.0275, 0.0355] \\
6 & Socio-Cultural & 0.8282 & 0.0157 & [0.0119, 0.0195] & 0.0345 & [0.0307, 0.0383] \\
7 & Grammar \& Function & 0.8733 & 0.0025 & [-0.0014, 0.0063] & 0.0279 & [0.0240, 0.0318] \\
7 & Socio-Cultural & 0.8282 & 0.0158 & [0.0120, 0.0196] & 0.0342 & [0.0304, 0.0380] \\
8 & Grammar \& Function & 0.8811 & 0.0062 & [0.0022, 0.0102] & 0.0271 & [0.0231, 0.0311] \\
8 & Socio-Cultural & 0.8317 & 0.0171 & [0.0133, 0.0209] & 0.0331 & [0.0293, 0.0369] \\
9 & Grammar \& Function & 0.8791 & 0.0091 & [0.0050, 0.0132] & 0.0278 & [0.0238, 0.0319] \\
9 & Socio-Cultural & 0.8365 & 0.0171 & [0.0134, 0.0208] & 0.0325 & [0.0289, 0.0362] \\
10 & Grammar \& Function & 0.8836 & 0.0090 & [0.0050, 0.0130] & 0.0272 & [0.0232, 0.0312] \\
10 & Socio-Cultural & 0.8508 & 0.0151 & [0.0117, 0.0185] & 0.0302 & [0.0268, 0.0336] \\
11 & Grammar \& Function & 0.8897 & 0.0071 & [0.0032, 0.0111] & 0.0261 & [0.0221, 0.0300] \\
11 & Socio-Cultural & 0.8622 & 0.0148 & [0.0116, 0.0180] & 0.0283 & [0.0251, 0.0315] \\
12 & Grammar \& Function & 0.8924 & 0.0057 & [0.0017, 0.0096] & 0.0285 & [0.0246, 0.0325] \\
12 & Socio-Cultural & 0.8742 & 0.0165 & [0.0136, 0.0195] & 0.0263 & [0.0233, 0.0292] \\
\end{longtable}
\endgroup

\begin{longtable}{lp{.43\textwidth}rr}
\caption{Frequency-adjusted item-level models for both encoders. Estimates are REML fixed effects; intervals are 95\% Wald intervals, not Bayesian credible intervals. Each fit uses all 635,040 observations.}\label{tab:frequency}\\
\toprule
Encoder & Predictor & Estimate & 95\% Wald interval\\\midrule\endfirsthead
\toprule
Encoder & Predictor & Estimate & 95\% Wald interval\\\midrule\endhead
\bottomrule\endfoot
mbert & Intercept & 0.93557 & [0.92232, 0.94881] \\
mbert & Geo\_Z & 0.00286 & [0.00221, 0.00351] \\
mbert & ASJP\_Z & 0.04151 & [0.04090, 0.04211] \\
mbert & Frequency\_Z & -0.02620 & [-0.03228, -0.02012] \\
mbert & DomainNature\_Motion & -0.00856 & [-0.02240, 0.00528] \\
mbert & DomainSocio\_Cultural & -0.00775 & [-0.02079, 0.00529] \\
mbert & DomainGrammar\_Function & -0.02010 & [-0.04058, 0.00038] \\
mbert & Geo\_Z:DomainNature\_Motion & -0.00127 & [-0.00216, -0.00038] \\
mbert & Geo\_Z:DomainSocio\_Cultural & 0.00295 & [0.00211, 0.00380] \\
mbert & Geo\_Z:DomainGrammar\_Function & 0.00898 & [0.00780, 0.01017] \\
mbert & ASJP\_Z:DomainNature\_Motion & 0.00389 & [0.00300, 0.00477] \\
mbert & ASJP\_Z:DomainSocio\_Cultural & -0.00395 & [-0.00479, -0.00310] \\
mbert & ASJP\_Z:DomainGrammar\_Function & 0.00310 & [0.00192, 0.00429] \\
xlmr & Intercept & 0.91213 & [0.89302, 0.93124] \\
xlmr & Geo\_Z & 0.00526 & [0.00391, 0.00662] \\
xlmr & ASJP\_Z & 0.03331 & [0.03205, 0.03457] \\
xlmr & Frequency\_Z & -0.01414 & [-0.02424, -0.00404] \\
xlmr & DomainNature\_Motion & 0.02264 & [-0.00037, 0.04565] \\
xlmr & DomainSocio\_Cultural & -0.00940 & [-0.03108, 0.01228] \\
xlmr & DomainGrammar\_Function & -0.04103 & [-0.07507, -0.00698] \\
xlmr & Geo\_Z:DomainNature\_Motion & -0.00204 & [-0.00390, -0.00019] \\
xlmr & Geo\_Z:DomainSocio\_Cultural & -0.00063 & [-0.00239, 0.00114] \\
xlmr & Geo\_Z:DomainGrammar\_Function & 0.00911 & [0.00664, 0.01158] \\
xlmr & ASJP\_Z:DomainNature\_Motion & 0.00083 & [-0.00103, 0.00269] \\
xlmr & ASJP\_Z:DomainSocio\_Cultural & -0.00251 & [-0.00427, -0.00075] \\
xlmr & ASJP\_Z:DomainGrammar\_Function & 0.00849 & [0.00602, 0.01096] \\
\end{longtable}

\begin{longtable}{rl}
\caption{The 81 languages retained after excluding degenerate Javanese geometry.}\label{tab:languages}\\
\toprule
Index & Language \\
\midrule
\endfirsthead
\toprule
Index & Language \\
\midrule
\endhead
\midrule
\multicolumn{2}{r}{Continued on next page} \\
\endfoot
\bottomrule
\endlastfoot
1 & Afrikaans \\
2 & Albanian \\
3 & Arabic \\
4 & Armenian \\
5 & Asturian \\
6 & Azerbaijani \\
7 & Bashkir \\
8 & Basque \\
9 & Belarusian \\
10 & Bengali \\
11 & Bokmål \\
12 & Breton \\
13 & Bulgarian \\
14 & Burmese \\
15 & Catalan \\
16 & Chechen \\
17 & Chuvash \\
18 & Czech \\
19 & Danish \\
20 & Dutch \\
21 & English \\
22 & Estonian \\
23 & Finnish \\
24 & French \\
25 & Galician \\
26 & Georgian \\
27 & German \\
28 & Greek \\
29 & Gujarati \\
30 & Hebrew \\
31 & Hindi \\
32 & Hungarian \\
33 & Icelandic \\
34 & Indonesian \\
35 & Italian \\
36 & Japanese \\
37 & Kannada \\
38 & Kazakh \\
39 & Korean \\
40 & Kyrgyz \\
41 & Latvian \\
42 & Lithuanian \\
43 & Low Saxon \\
44 & Macedonian \\
45 & Malagasy \\
46 & Malay \\
47 & Malayalam \\
48 & Mandarin \\
49 & Marathi \\
50 & Mongolian \\
51 & Nepali \\
52 & Nynorsk \\
53 & Occitan \\
54 & Persian \\
55 & Piedmontese \\
56 & Polish \\
57 & Portuguese \\
58 & Punjabi \\
59 & Romanian \\
60 & Russian \\
61 & Scots \\
62 & Serbo-Croatian \\
63 & Slovak \\
64 & Slovene \\
65 & Spanish \\
66 & Sundanese \\
67 & Swahili \\
68 & Swedish \\
69 & Tagalog \\
70 & Tajik \\
71 & Tamil \\
72 & Tatar \\
73 & Telugu \\
74 & Thai \\
75 & Turkish \\
76 & Ukrainian \\
77 & Urdu \\
78 & Uzbek \\
79 & Vietnamese \\
80 & Welsh \\
81 & West Frisian \\
\end{longtable}

\begin{longtable}{p{.34\textwidth}p{.56\textwidth}}
\caption{XLM-R-large extraction and fixed-sample RSA provenance. The checkpoint was loaded from local server storage without network fallback.}\label{tab:cross-extraction}\\
\toprule
Item & Value \\
\midrule
\endfirsthead
\toprule
Item & Value \\
\midrule
\endhead
\midrule
\multicolumn{2}{r}{Continued on next page} \\
\endfoot
\bottomrule
\endlastfoot
Checkpoint & \texttt{xlm-roberta-large} \\
Local checkpoint directory & \texttt{models/xlm-roberta-large} \\
Architecture & XLMRobertaModel \\
Layers / hidden size & 24 / 1024 \\
Layer / pooling & final; mean non-special-token states within alternative; mean across comma-delimited alternatives \\
Input & isolated unmarked lexical forms \\
Languages completed & 81 / 81 \\
Completed QC rows & 81 \\
Software & Python 3.12.4; torch 2.4.0+cu121; transformers 5.14.1 \\
Config SHA-256 & \texttt{ec7c3a99\allowbreak{}c58a38eb\allowbreak{}b702a229\allowbreak{}7f1c4586\allowbreak{}944e841b\allowbreak{}5eb29d20\allowbreak{}1ff87e0f\allowbreak{}9c9abe0e} \\
XLM-R RSA SHA-256 & \texttt{dee33af4\allowbreak{}d8481b33\allowbreak{}c34995b2\allowbreak{}91c5c47a\allowbreak{}67359811\allowbreak{}02f6f687\allowbreak{}0a0a46c6\allowbreak{}7e5f4420} \\
\end{longtable}

\begin{longtable}{lllp{.12\textwidth}p{.19\textwidth}}
\caption{Fixed-sample domain means and leave-one-language-out ranges for both encoders and both taxonomies. All rows use the same 81 languages, 196 concepts, and 3,240 dyads.}\label{tab:cross-domain}\\
\toprule
Encoder & Taxonomy & Domain & Mean & LOO range \\
\midrule
\endfirsthead
\toprule
Encoder & Taxonomy & Domain & Mean & LOO range \\
\midrule
\endhead
\midrule
\multicolumn{5}{r}{Continued on next page} \\
\endfoot
\bottomrule
\endlastfoot
mBERT & Primary & Biological Core & 0.900 & [0.898, 0.904] \\
XLM-R-large & Primary & Biological Core & 0.932 & [0.930, 0.933] \\
mBERT & Primary & Grammar \& Function & 0.892 & [0.889, 0.895] \\
XLM-R-large & Primary & Grammar \& Function & 0.967 & [0.966, 0.969] \\
mBERT & Primary & Nature \& Motion & 0.924 & [0.923, 0.927] \\
XLM-R-large & Primary & Nature \& Motion & 0.977 & [0.976, 0.978] \\
mBERT & Primary & Socio-Cultural & 0.874 & [0.871, 0.877] \\
XLM-R-large & Primary & Socio-Cultural & 0.922 & [0.920, 0.924] \\
mBERT & External & Biological Core & 0.918 & [0.916, 0.920] \\
XLM-R-large & External & Biological Core & 0.953 & [0.952, 0.954] \\
mBERT & External & Grammar \& Function & 0.892 & [0.889, 0.894] \\
XLM-R-large & External & Grammar \& Function & 0.963 & [0.962, 0.965] \\
mBERT & External & Nature \& Motion & 0.921 & [0.919, 0.923] \\
XLM-R-large & External & Nature \& Motion & 0.973 & [0.972, 0.974] \\
mBERT & External & Socio-Cultural & 0.883 & [0.880, 0.886] \\
XLM-R-large & External & Socio-Cultural & 0.938 & [0.937, 0.940] \\
\end{longtable}

\begin{longtable}{llrrr}
\caption{Dyad-level agreement between mBERT and XLM-R-large RSA dissimilarities. Median agreement is the fraction of dyads assigned to the same side of each model's median.}\label{tab:cross-agreement}\\
\toprule
Taxonomy & Domain & $\rho$ & Median agree. & Mean abs. diff. \\
\midrule
\endfirsthead
\toprule
Taxonomy & Domain & $\rho$ & Median agree. & Mean abs. diff. \\
\midrule
\endhead
\midrule
\multicolumn{5}{r}{Continued on next page} \\
\endfoot
\bottomrule
\endlastfoot
Primary & All concepts & 0.569 & 0.685 & 0.046 \\
Primary & Biological Core & 0.361 & 0.622 & 0.068 \\
Primary & Grammar \& Function & 0.169 & 0.553 & 0.114 \\
Primary & Nature \& Motion & 0.209 & 0.559 & 0.069 \\
Primary & Socio-Cultural & 0.531 & 0.680 & 0.075 \\
External & All concepts & 0.569 & 0.685 & 0.046 \\
External & Biological Core & 0.275 & 0.593 & 0.066 \\
External & Grammar \& Function & 0.238 & 0.575 & 0.095 \\
External & Nature \& Motion & 0.270 & 0.588 & 0.068 \\
External & Socio-Cultural & 0.522 & 0.674 & 0.078 \\
\end{longtable}

\begin{longtable}{llp{.38\textwidth}cc}
\caption{Leave-one-language-out stability and model-specific domain ordering. Each model's observed ordering (shown with abbreviated domain labels) was preserved in all 81 leave-one-out subsamples under both taxonomies.}\label{tab:cross-order}\\
\toprule
Encoder & Taxonomy & Observed domain order & Shared extremes & Full order \\
\midrule
\endfirsthead
\toprule
Encoder & Taxonomy & Observed domain order & Shared extremes & Full order \\
\midrule
\endhead
\midrule
\multicolumn{5}{r}{Continued on next page} \\
\endfoot
\bottomrule
\endlastfoot
XLM-R-large & Primary & S-Cultural $<$ Bio. Core $<$ Grammar $<$ Nature & 81/81 & 81/81 \\
XLM-R-large & External & S-Cultural $<$ Bio. Core $<$ Grammar $<$ Nature & 81/81 & 81/81 \\
mBERT & Primary & S-Cultural $<$ Grammar $<$ Bio. Core $<$ Nature & 81/81 & 81/81 \\
mBERT & External & S-Cultural $<$ Grammar $<$ Bio. Core $<$ Nature & 81/81 & 81/81 \\
\end{longtable}

\begin{longtable}{p{.30\textwidth}p{.60\textwidth}}
\caption{Software and provenance metadata for the analysis. Heavy numerical analyses were executed on the remote server; no local Bayesian sampling or model inference was used.}\label{tab:provenance}\\
\toprule
Item & Value \\
\midrule
\endfirsthead
\toprule
Item & Value \\
\midrule
\endhead
\midrule
\multicolumn{2}{r}{Continued on next page} \\
\endfoot
\bottomrule
\endlastfoot
Encoder & mBERT and XLM-R-large; separate primary/external fits plus pooled model \\
Python & 3.12.4 \\
NumPy / pandas & 1.26.4 / 2.3.3 \\
PyTorch / Transformers & 2.4.0+cu121 / 5.14.1 \\
CmdStan & 2.38.0; invoked by the remote Python driver \\
Chains & 4 \\
Warm-up / sampling & Full-support runs: 3000 / 3000 per chain; no reduced-iteration fit enters manuscript claims \\
Seed & 42 \\
mBERT RSA input SHA-256 & \texttt{38563c1d\allowbreak{}16486c0d\allowbreak{}664381aa\allowbreak{}39c04b6b\allowbreak{}eb1184b3\allowbreak{}719b4caa\allowbreak{}0cc1041a\allowbreak{}2de4ad9b} \\
XLM-R RSA input SHA-256 & \texttt{dee33af4\allowbreak{}d8481b33\allowbreak{}c34995b2\allowbreak{}91c5c47a\allowbreak{}67359811\allowbreak{}02f6f687\allowbreak{}0a0a46c6\allowbreak{}7e5f4420} \\
Geographic matrix SHA-256 & \texttt{3fe9fd1e\allowbreak{}bf3a4a19\allowbreak{}9d5671b0\allowbreak{}92944b0e\allowbreak{}9feff229\allowbreak{}2a1c00d0\allowbreak{}ad872502\allowbreak{}4b6c3862} \\
ASJP matrix SHA-256 & \texttt{8ecfe162\allowbreak{}63f77cbd\allowbreak{}c894aeac\allowbreak{}a5b18c77\allowbreak{}50593fbd\allowbreak{}8c7716b4\allowbreak{}8de5a904\allowbreak{}18521c9a} \\
Language-order file SHA-256 & \texttt{d8812813\allowbreak{}78d293d3\allowbreak{}38ca8b17\allowbreak{}c9a840a8\allowbreak{}fd87002e\allowbreak{}324867f6\allowbreak{}99813a95\allowbreak{}429b9fbf} \\
Taxonomy crosswalk SHA-256 & \texttt{2661fed4\allowbreak{}24363e67\allowbreak{}9a4f3cede\allowbreak{}9949f1cb\allowbreak{}ac664f04\allowbreak{}19a6d302\allowbreak{}f5abb820\allowbreak{}da70f29} \\
Global Stan SHA-256 & \texttt{d53b6d6d\allowbreak{}cd783fc4\allowbreak{}feeadd5a\allowbreak{}82cfdf27\allowbreak{}847da9d1\allowbreak{}df23b9a1\allowbreak{}be6871a8\allowbreak{}e65d8f14} \\
Domain Stan SHA-256 & \texttt{8762ddb5\allowbreak{}d8fb9e16\allowbreak{}580c61c2\allowbreak{}5286b816\allowbreak{}27dd0c16\allowbreak{}8a234b7d\allowbreak{}1c391b72\allowbreak{}3a55da3e} \\
Pooled Stan SHA-256 & \texttt{dce7806f\allowbreak{}c13131d2\allowbreak{}7551f523\allowbreak{}541ffaac\allowbreak{}712f2eb1\allowbreak{}7822b531\allowbreak{}532d6d3f\allowbreak{}b5e89ac2} \\
\end{longtable}

\begin{longtable}{r p{.13\textwidth} p{.16\textwidth} p{.16\textwidth} p{.25\textwidth} l}
\caption{Complete mapping of the 207 Wiktionary-numbered study concepts. The external mapping uses a study-specific alignment to Bowern/Concepticon entries, linked WOLD/IDS fields, WordNet 3.0 lexnames where applicable, and explicit functional rules. The auxiliary alignment does not imply identity of the original concept lists; the frozen WOLD/IDS-informed macro-domain rule is retained separately from official source fields.}\label{tab:mapping}\\
\toprule
ID & Concept & Primary & External & External field & Confidence \\
\midrule
\endfirsthead
\toprule
ID & Concept & Primary & External & External field & Confidence \\
\midrule
\endhead
\midrule
\multicolumn{6}{r}{Continued on next page} \\
\endfoot
\bottomrule
\endlastfoot
1 & I & Biological Core & Grammar \& Function & Pronouns & high \\
2 & you & Biological Core & Grammar \& Function & Pronouns & high \\
3 & he,she,it & Biological Core & Grammar \& Function & Pronouns & medium \\
4 & we & Biological Core & Grammar \& Function & Pronouns & high \\
5 & you & Biological Core & Grammar \& Function & Pronouns & medium \\
6 & they & Biological Core & Grammar \& Function & Pronouns & high \\
7 & this & Grammar \& Function & Grammar \& Function & Deixis and reference & high \\
8 & that & Grammar \& Function & Grammar \& Function & Deixis and reference & high \\
9 & here & Grammar \& Function & Grammar \& Function & Space and location & high \\
10 & there & Grammar \& Function & Grammar \& Function & Space and location & high \\
11 & who & Grammar \& Function & Grammar \& Function & Function words & high \\
12 & what & Grammar \& Function & Grammar \& Function & Function words & high \\
13 & where & Grammar \& Function & Grammar \& Function & Function words & high \\
14 & when & Grammar \& Function & Grammar \& Function & Function words & high \\
15 & how & Grammar \& Function & Grammar \& Function & Function words & high \\
16 & not & Grammar \& Function & Grammar \& Function & Function words or residual field & high \\
17 & all & Grammar \& Function & Grammar \& Function & Quantity & high \\
18 & many & Grammar \& Function & Grammar \& Function & Quantity & high \\
19 & some & Grammar \& Function & Grammar \& Function & Quantity & high \\
20 & few & Grammar \& Function & Grammar \& Function & Quantity & high \\
21 & other & Grammar \& Function & Grammar \& Function & Function words or residual field & high \\
22 & one & Socio-Cultural & Grammar \& Function & Quantity & high \\
23 & two & Socio-Cultural & Grammar \& Function & Quantity & high \\
24 & three & Socio-Cultural & Grammar \& Function & Quantity & high \\
25 & four & Socio-Cultural & Grammar \& Function & Quantity & high \\
26 & five & Socio-Cultural & Socio-Cultural & Quantity & high \\
27 & big & Socio-Cultural & Socio-Cultural & Dimension & high \\
28 & long & Socio-Cultural & Socio-Cultural & Dimension & high \\
29 & wide & Socio-Cultural & Socio-Cultural & Dimension & high \\
30 & thick & Socio-Cultural & Socio-Cultural & Dimension & high \\
31 & heavy & Socio-Cultural & Socio-Cultural & Attributes and qualities & high \\
32 & small & Socio-Cultural & Socio-Cultural & Dimension & high \\
33 & short & Socio-Cultural & Socio-Cultural & Dimension & high \\
34 & narrow & Socio-Cultural & Socio-Cultural & Dimension & high \\
35 & thin & Socio-Cultural & Socio-Cultural & Dimension & high \\
36 & woman & Biological Core & Biological Core & People & high \\
37 & man & Biological Core & Biological Core & People & medium \\
38 & man & Biological Core & Biological Core & People & medium \\
39 & child & Biological Core & Biological Core & Kinship & high \\
40 & wife & Biological Core & Biological Core & Kinship & high \\
41 & husband & Biological Core & Biological Core & Kinship & high \\
42 & mother & Biological Core & Biological Core & Kinship & high \\
43 & father & Biological Core & Biological Core & Kinship & high \\
44 & animal & Nature \& Motion & Nature \& Motion & Animals & high \\
45 & fish & Nature \& Motion & Nature \& Motion & Animals & high \\
46 & bird & Nature \& Motion & Nature \& Motion & Animals & high \\
47 & dog & Nature \& Motion & Nature \& Motion & Animals & high \\
48 & louse & Nature \& Motion & Nature \& Motion & Animals & high \\
49 & snake & Nature \& Motion & Nature \& Motion & Animals & high \\
50 & worm & Nature \& Motion & Nature \& Motion & Animals & high \\
51 & tree & Nature \& Motion & Nature \& Motion & Plants & high \\
52 & forest & Nature \& Motion & Nature \& Motion & Plants & high \\
53 & stick & Nature \& Motion & Nature \& Motion & Plants & high \\
54 & fruit & Nature \& Motion & Nature \& Motion & Plants & high \\
55 & seed & Nature \& Motion & Nature \& Motion & Plants & high \\
56 & leaf & Nature \& Motion & Nature \& Motion & Plants & high \\
57 & root & Nature \& Motion & Nature \& Motion & Plants & high \\
58 & bark & Nature \& Motion & Nature \& Motion & Plants & high \\
59 & flower & Nature \& Motion & Nature \& Motion & Plants & high \\
60 & grass & Nature \& Motion & Nature \& Motion & Plants & high \\
61 & rope & Socio-Cultural & Socio-Cultural & Material culture & high \\
62 & skin & Biological Core & Biological Core & Body and substance & high \\
63 & meat & Biological Core & Biological Core & Body and substance & high \\
64 & blood & Biological Core & Biological Core & Body and substance & high \\
65 & bone & Biological Core & Biological Core & Body and substance & high \\
66 & fat & Biological Core & Biological Core & Body and substance & high \\
67 & egg & Biological Core & Biological Core & Body and substance & high \\
68 & horn & Biological Core & Biological Core & Body and substance & high \\
69 & tail & Biological Core & Biological Core & Body and substance & high \\
70 & feather & Biological Core & Biological Core & Body and substance & high \\
71 & hair & Biological Core & Biological Core & Body and substance & high \\
72 & head & Biological Core & Biological Core & Body and substance & high \\
73 & ear & Biological Core & Biological Core & Body and substance & high \\
74 & eye & Biological Core & Biological Core & Body and substance & high \\
75 & nose & Biological Core & Biological Core & Body and substance & high \\
76 & mouth & Biological Core & Biological Core & Body and substance & high \\
77 & tooth & Biological Core & Biological Core & Body and substance & high \\
78 & tongue & Biological Core & Biological Core & Body and substance & high \\
79 & fingernail & Biological Core & Biological Core & Body and substance & high \\
80 & foot & Biological Core & Biological Core & Body and substance & high \\
81 & leg & Biological Core & Biological Core & Body and substance & high \\
82 & knee & Biological Core & Biological Core & Body and substance & high \\
83 & hand & Biological Core & Biological Core & Body and substance & high \\
84 & wing & Biological Core & Biological Core & Body and substance & high \\
85 & belly & Biological Core & Biological Core & Body and substance & high \\
86 & guts & Biological Core & Biological Core & Body and substance & high \\
87 & neck & Biological Core & Biological Core & Body and substance & high \\
88 & back & Biological Core & Biological Core & Body and substance & high \\
89 & breast & Biological Core & Biological Core & Body and substance & high \\
90 & heart & Biological Core & Biological Core & Body and substance & high \\
91 & liver & Biological Core & Biological Core & Body and substance & high \\
92 & drink & Biological Core & Biological Core & Food, bodily action, and speech & high \\
93 & eat & Biological Core & Biological Core & Food, bodily action, and speech & high \\
94 & bite & Biological Core & Biological Core & Food, bodily action, and speech & high \\
95 & suck & Biological Core & Biological Core & Food, bodily action, and speech & high \\
96 & spit & Biological Core & Biological Core & Food, bodily action, and speech & high \\
97 & vomit & Biological Core & Biological Core & Food, bodily action, and speech & high \\
98 & blow & Biological Core & Biological Core & Food, bodily action, and speech & high \\
99 & breathe & Biological Core & Biological Core & Body and sensation & high \\
100 & laugh & Biological Core & Biological Core & Body and sensation & high \\
101 & see & Biological Core & Biological Core & Body and sensation & high \\
102 & hear & Biological Core & Biological Core & Body and sensation & high \\
103 & know & Biological Core & Socio-Cultural & Body and sensation & medium \\
104 & think & Biological Core & Socio-Cultural & Body and sensation & medium \\
105 & smell & Biological Core & Biological Core & Body and sensation & high \\
106 & fear & Biological Core & Socio-Cultural & Body and sensation & medium \\
107 & sleep & Biological Core & Biological Core & Body and sensation & high \\
108 & live & Biological Core & Biological Core & Body and sensation & high \\
109 & die & Biological Core & Biological Core & Body and sensation & high \\
110 & kill & Socio-Cultural & Socio-Cultural & Action and manipulation & high \\
111 & fight & Socio-Cultural & Socio-Cultural & Action and manipulation & high \\
112 & hunt & Nature \& Motion & Socio-Cultural & Material culture & medium \\
113 & hit & Socio-Cultural & Socio-Cultural & Action and manipulation & high \\
114 & cut & Socio-Cultural & Socio-Cultural & Action and manipulation & high \\
115 & split & Socio-Cultural & Socio-Cultural & Action and manipulation & high \\
116 & stab & Socio-Cultural & Socio-Cultural & Action and manipulation & high \\
117 & scratch & Socio-Cultural & Socio-Cultural & Action and manipulation & high \\
118 & dig & Socio-Cultural & Socio-Cultural & Action and manipulation & high \\
119 & swim & Nature \& Motion & Nature \& Motion & Space and motion & high \\
120 & fly & Nature \& Motion & Nature \& Motion & Space and motion & high \\
121 & walk & Nature \& Motion & Nature \& Motion & Space and motion & high \\
122 & come & Nature \& Motion & Nature \& Motion & Space and motion & high \\
123 & lie & Nature \& Motion & Nature \& Motion & Space and motion & high \\
124 & sit & Nature \& Motion & Nature \& Motion & Space and motion & high \\
125 & stand & Nature \& Motion & Nature \& Motion & Space and motion & high \\
126 & turn & Nature \& Motion & Nature \& Motion & Space and motion & high \\
127 & fall & Nature \& Motion & Nature \& Motion & Space and motion & high \\
128 & give & Socio-Cultural & Socio-Cultural & Action and manipulation & high \\
129 & hold & Socio-Cultural & Socio-Cultural & Action and manipulation & high \\
130 & squeeze & Socio-Cultural & Socio-Cultural & Action and manipulation & high \\
131 & rub & Socio-Cultural & Socio-Cultural & Action and manipulation & high \\
132 & wash & Socio-Cultural & Socio-Cultural & Action and manipulation & high \\
133 & wipe & Socio-Cultural & Socio-Cultural & Action and manipulation & high \\
134 & pull & Socio-Cultural & Socio-Cultural & Action and manipulation & high \\
135 & push & Socio-Cultural & Socio-Cultural & Action and manipulation & high \\
136 & throw & Socio-Cultural & Socio-Cultural & Action and manipulation & high \\
137 & tie & Socio-Cultural & Socio-Cultural & Action and manipulation & high \\
138 & sew & Socio-Cultural & Socio-Cultural & Action and manipulation & high \\
139 & count & Biological Core & Grammar \& Function & Food, bodily action, and speech & medium \\
140 & say & Biological Core & Socio-Cultural & Food, bodily action, and speech & medium \\
141 & sing & Biological Core & Socio-Cultural & Food, bodily action, and speech & medium \\
142 & play & Nature \& Motion & Socio-Cultural & Function words or residual field & medium \\
143 & float & Nature \& Motion & Nature \& Motion & Space and motion & high \\
144 & flow & Nature \& Motion & Nature \& Motion & Space and motion & high \\
145 & freeze & Biological Core & Nature \& Motion & Body and sensation & high \\
146 & swell & Biological Core & Nature \& Motion & Body and sensation & high \\
147 & sun & Nature \& Motion & Nature \& Motion & Natural environment & high \\
148 & moon & Nature \& Motion & Nature \& Motion & Natural environment & high \\
149 & star & Nature \& Motion & Nature \& Motion & Natural environment & high \\
150 & water & Nature \& Motion & Nature \& Motion & Natural environment & high \\
151 & rain & Nature \& Motion & Nature \& Motion & Natural environment & high \\
152 & river & Nature \& Motion & Nature \& Motion & Natural environment & high \\
153 & lake & Nature \& Motion & Nature \& Motion & Natural environment & high \\
154 & sea & Nature \& Motion & Nature \& Motion & Natural environment & high \\
155 & salt & Nature \& Motion & Nature \& Motion & Natural environment & high \\
156 & stone & Nature \& Motion & Nature \& Motion & Natural environment & high \\
157 & sand & Nature \& Motion & Nature \& Motion & Natural environment & high \\
158 & dust & Nature \& Motion & Nature \& Motion & Natural environment & high \\
159 & earth & Nature \& Motion & Nature \& Motion & Natural environment & high \\
160 & cloud & Nature \& Motion & Nature \& Motion & Natural environment & high \\
161 & fog & Nature \& Motion & Nature \& Motion & Natural environment & high \\
162 & sky & Nature \& Motion & Nature \& Motion & Natural environment & high \\
163 & wind & Nature \& Motion & Nature \& Motion & Natural environment & high \\
164 & snow & Nature \& Motion & Nature \& Motion & Natural environment & high \\
165 & ice & Nature \& Motion & Nature \& Motion & Natural environment & high \\
166 & smoke & Nature \& Motion & Nature \& Motion & Natural environment & high \\
167 & fire & Nature \& Motion & Nature \& Motion & Natural environment & high \\
168 & ash & Nature \& Motion & Nature \& Motion & Natural environment & high \\
169 & burn & Biological Core & Nature \& Motion & Body and sensation & medium \\
170 & road & Socio-Cultural & Socio-Cultural & Material culture & high \\
171 & mountain & Nature \& Motion & Nature \& Motion & Natural environment & high \\
172 & red & Socio-Cultural & Socio-Cultural & Colour & high \\
173 & green & Socio-Cultural & Socio-Cultural & Colour & high \\
174 & yellow & Socio-Cultural & Socio-Cultural & Colour & high \\
175 & white & Socio-Cultural & Socio-Cultural & Colour & high \\
176 & black & Socio-Cultural & Socio-Cultural & Colour & high \\
177 & night & Socio-Cultural & Socio-Cultural & Time & high \\
178 & day & Socio-Cultural & Socio-Cultural & Time & high \\
179 & year & Socio-Cultural & Socio-Cultural & Time & high \\
180 & warm & Socio-Cultural & Socio-Cultural & Attributes and qualities & high \\
181 & cold & Socio-Cultural & Socio-Cultural & Attributes and qualities & high \\
182 & full & Socio-Cultural & Socio-Cultural & Attributes and qualities & high \\
183 & new & Socio-Cultural & Socio-Cultural & Attributes and qualities & high \\
184 & old & Socio-Cultural & Socio-Cultural & Attributes and qualities & high \\
185 & good & Socio-Cultural & Socio-Cultural & Attributes and qualities & high \\
186 & bad & Socio-Cultural & Socio-Cultural & Attributes and qualities & high \\
187 & rotten & Socio-Cultural & Socio-Cultural & Attributes and qualities & high \\
188 & dirty & Socio-Cultural & Socio-Cultural & Attributes and qualities & high \\
189 & straight & Socio-Cultural & Socio-Cultural & Attributes and qualities & high \\
190 & round & Socio-Cultural & Socio-Cultural & Attributes and qualities & high \\
191 & sharp & Socio-Cultural & Socio-Cultural & Attributes and qualities & high \\
192 & dull & Socio-Cultural & Socio-Cultural & Attributes and qualities & high \\
193 & smooth & Socio-Cultural & Socio-Cultural & Attributes and qualities & high \\
194 & wet & Socio-Cultural & Socio-Cultural & Attributes and qualities & high \\
195 & dry & Socio-Cultural & Socio-Cultural & Attributes and qualities & high \\
196 & correct & Socio-Cultural & Socio-Cultural & Attributes and qualities & high \\
197 & near & Grammar \& Function & Grammar \& Function & Space and location & high \\
198 & far & Grammar \& Function & Grammar \& Function & Space and location & high \\
199 & right & Grammar \& Function & Grammar \& Function & Space and location & medium \\
200 & left & Grammar \& Function & Grammar \& Function & Space and location & high \\
201 & at & Grammar \& Function & Grammar \& Function & Space and location & high \\
202 & in & Grammar \& Function & Grammar \& Function & Space and location & high \\
203 & with & Grammar \& Function & Grammar \& Function & Space and location & high \\
204 & and & Grammar \& Function & Grammar \& Function & Function words or residual field & high \\
205 & if & Grammar \& Function & Grammar \& Function & Function words or residual field & high \\
206 & because & Grammar \& Function & Grammar \& Function & Function words or residual field & high \\
207 & name & Grammar \& Function & Socio-Cultural & Function words or residual field & medium \\
\end{longtable}

\endgroup
\begin{longtable}{p{.10\textwidth}p{.23\textwidth}p{.30\textwidth}p{.26\textwidth}}
\caption{Complete pooled two-domain contrasts under default and sensitivity priors. Within-encoder and between-encoder estimands are listed separately.}\label{tab:pooled_complete}\\
\toprule
Mapping & Prior branch & Estimand & Mean [95\% CrI]\\\midrule\endfirsthead
\toprule
Mapping & Prior branch & Estimand & Mean [95\% CrI]\\\midrule\endhead
\bottomrule\endfoot
primary & Default (0.2) & Nature \& Motion - Socio-Cultural | mBERT & 0.05031 [0.04702, 0.05366] \\
primary & Default (0.2) & Nature \& Motion - Socio-Cultural | XLM-R-large & 0.05472 [0.05128, 0.05803] \\
primary & Default (0.2) & (XLM-R-large - mBERT) endpoint contrast & 0.00441 [-0.00034, 0.00897] \\
primary & Sensitivity (0.5) & Nature \& Motion - Socio-Cultural | mBERT & 0.05030 [0.04695, 0.05369] \\
primary & Sensitivity (0.5) & Nature \& Motion - Socio-Cultural | XLM-R-large & 0.05469 [0.05129, 0.05805] \\
primary & Sensitivity (0.5) & (XLM-R-large - mBERT) endpoint contrast & 0.00439 [-0.00046, 0.00915] \\
external & Default (0.2) & Nature \& Motion - Socio-Cultural | mBERT & 0.03749 [0.03454, 0.04052] \\
external & Default (0.2) & Nature \& Motion - Socio-Cultural | XLM-R-large & 0.03461 [0.03163, 0.03758] \\
external & Default (0.2) & (XLM-R-large - mBERT) endpoint contrast & -0.00288 [-0.00709, 0.00138] \\
external & Sensitivity (0.5) & Nature \& Motion - Socio-Cultural | mBERT & 0.03750 [0.03444, 0.04047] \\
external & Sensitivity (0.5) & Nature \& Motion - Socio-Cultural | XLM-R-large & 0.03461 [0.03157, 0.03768] \\
external & Sensitivity (0.5) & (XLM-R-large - mBERT) endpoint contrast & -0.00289 [-0.00709, 0.00139] \\
\end{longtable}
\clearpage
\section{Encoder comparison details}
Both encoders use the same 81 languages, 196 complete-case concept identifiers, domain mappings, lexical input rule, final-layer nonspecial-token pooling, within-set centering, RSA construction, and predictor matrices. The domain fits contain 12,960 rows each. The pooled model contains 25,920 rows. The mBERT extraction manifest records 83 requested extraction files, while the assembled study inventory contained 82 languages and the comparative analysis retained 81; extraction requests and analytic eligibility counts are different records.

The frequency-adjusted models use all 635,040 concept--dyad observations, English Zipf frequency, and exact Gaussian REML with marginalized language and concept effects. Their optimizer diagnostics report successful convergence and a full-rank 13-column fixed-effect design. These models operate on item-level cosine distances.

Equal-size subsampling and the 13-layer table use archived mBERT vectors identified in their respective manifests and serve as preliminary sensitivity analyses. The source taxonomy review forms document operational classifications.

\section{Supplementary figures}
\begin{figure}[!htbp]
\centering\includegraphics[width=\textwidth]{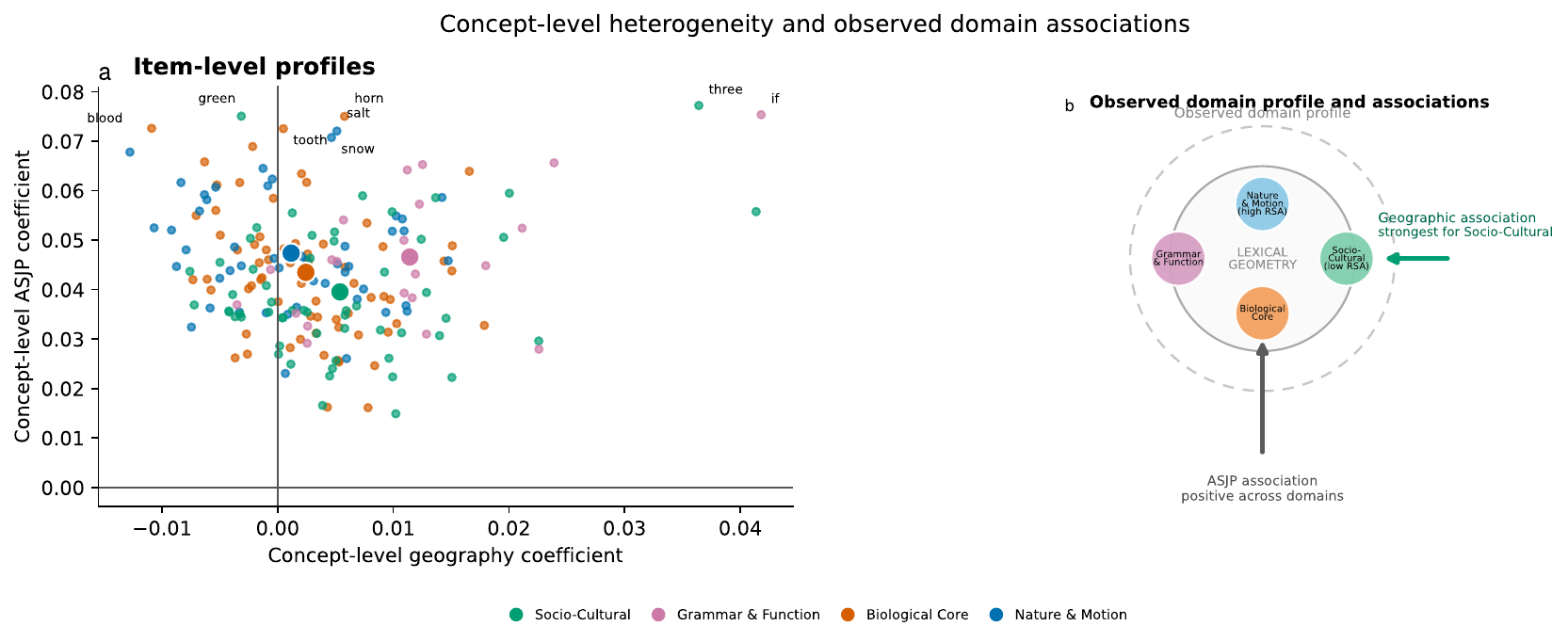}
\caption{\textbf{mBERT concept-level heterogeneity.} Panel (a) reports item-level regression coefficients for individual concepts. Panel (b) summarizes the full-inventory mBERT representational profile; arrows indicate observed associations. The geographic contrast shown is specific to the mBERT primary-taxonomy configuration.}\label{fig:conceptprofiles}
\end{figure}

%% file: references.bib
@book{murphy2002concepts,
  author = {Murphy, Gregory L.},
  title = {The Big Book of Concepts},
  year = {2002},
  publisher = {MIT Press},
  isbn = {9780262134095},
  url = {https://mitpress.mit.edu/9780262632997/the-big-book-of-concepts/}
}

@article{barsalou2008grounded,
  title = {Grounded cognition},
  author = {Barsalou, Lawrence W.},
  journal = {Annual Review of Psychology},
  volume = {59},
  pages = {617--645},
  year = {2008},
  doi = {10.1146/annurev.psych.59.103006.093639}
}

@article{spelke2007core,
  title = {Core knowledge},
  author = {Spelke, Elizabeth S. and Kinzler, Katherine D.},
  journal = {Developmental Science},
  volume = {10},
  number = {1},
  pages = {89--96},
  year = {2007},
  doi = {10.1111/j.1467-7687.2007.00569.x}
}

@article{evans2009myth,
  title={The myth of language universals: Language diversity and its importance for cognitive science},
  author={Evans, Nicholas and Levinson, Stephen C},
  journal={Behavioral and Brain Sciences},
  volume={32},
  number={5},
  pages={429--448},
  year={2009},
  publisher={Cambridge University Press},
  doi={10.1017/S0140525X0999094X}
}

@book{lucy1992language,
  title={Language diversity and thought: A reformulation of the linguistic relativity hypothesis},
  author={Lucy, John A},
  year={1992},
  publisher={Cambridge University Press},
  doi={10.1017/CBO9780511620843}
}

@article{wolff2011linguistic,
  title={Linguistic relativity},
  author={Wolff, Phillip and Holmes, Kevin J},
  journal={Wiley Interdisciplinary Reviews: Cognitive Science},
  volume={2},
  number={3},
  pages={253--265},
  year={2011},
  publisher={Wiley Online Library},
  doi={10.1002/wcs.104}
}

@article{malt2013thought,
  title={How thought is mapped into words},
  author={Malt, Barbara C and Majid, Asifa},
  journal={Wiley Interdisciplinary Reviews: Cognitive Science},
  volume={4},
  number={6},
  pages={583--597},
  year={2013},
  publisher={Wiley},
  doi={10.1002/wcs.1251}
}

@article{malt_2008,
  author = {Malt, Barbara C. and Gennari, Silvia and Imai, Mutsumi and Ameel, Eef and Tsuda, Naoaki and Majid, Asifa},
  title = {Talking About Walking: Biomechanics and the Language of Locomotion},
  journal = {Psychological Science},
  year = {2008},
  volume = {19},
  number = {3},
  pages = {232--240},
  doi = {10.1111/j.1467-9280.2008.02074.x},
  url = {https://doi.org/10.1111/j.1467-9280.2008.02074.x}
}

@article{dolscheid_2013,
  author = {Dolscheid, Sarah and Shayan, Shakila and Majid, Asifa and Casasanto, Daniel},
  title = {The Thickness of Musical Pitch: Psychophysical Evidence for Linguistic Relativity},
  journal = {Psychological Science},
  year = {2013},
  volume = {24},
  number = {5},
  pages = {613--621},
  doi = {10.1177/0956797612457374},
  url = {https://doi.org/10.1177/0956797612457374}
}

@article{dolscheid_2014,
  author = {Dolscheid, Sarah and Hunnius, Sabine and Casasanto, Daniel and Majid, Asifa},
  title = {Prelinguistic Infants Are Sensitive to Space-Pitch Associations Found Across Cultures},
  journal = {Psychological Science},
  year = {2014},
  volume = {25},
  number = {6},
  pages = {1256--1261},
  doi = {10.1177/0956797614528521},
  url = {https://doi.org/10.1177/0956797614528521}
}

@article{gibson2017color,
  title={Color naming across languages reflects color use},
  author={Gibson, Edward and Futrell, Richard and Jara-Ettinger, Julian and Mahowald, Kyle and Bergen, Leon and Ratnasingam, Sivalogeswaran and Gibson, Mitchell and Piantadosi, Steven T and Conway, Bevil R},
  journal={Proceedings of the National Academy of Sciences},
  volume={114},
  number={40},
  pages={10785--10790},
  year={2017},
  publisher={National Acad Sciences},
  doi={10.1073/pnas.1619666114}
}

@article{kemp2012kinship,
  title={Kinship categories across languages reflect general communicative principles},
  author={Kemp, Charles and Regier, Terry},
  journal={Science},
  volume={336},
  number={6084},
  pages={1049--1054},
  year={2012},
  publisher={American Association for the Advancement of Science},
  doi={10.1126/science.1218811}
}

@article{burenhult2008language,
  title={Language and landscape: A cross-linguistic perspective},
  author={Burenhult, Niclas and Levinson, Stephen C},
  journal={Language Sciences},
  volume={30},
  number={2-3},
  pages={135--150},
  year={2008},
  publisher={Elsevier},
  doi={10.1016/j.langsci.2006.12.028}
}

@article{regier2016languages,
  title={Languages support efficient communication about the environment: Words for snow revisited},
  author={Regier, Terry and Carstensen, Alexandra and Kemp, Charles},
  journal={PLoS One},
  volume={11},
  number={4},
  pages={e0151138},
  year={2016},
  publisher={Public Library of Science San Francisco, CA USA},
  doi={10.1371/journal.pone.0151138}
}

@article{youn2016universal,
  title={On the universal structure of human lexical semantics},
  author={Youn, Hyejin and Sutton, Logan and Smith, Eric and Moore, Cristopher and Wilkins, Jon F and Maddieson, Ian and Croft, William and Bhattacharya, Tanmoy},
  journal={Proceedings of the National Academy of Sciences},
  volume={113},
  number={7},
  pages={1766--1771},
  year={2016},
  publisher={National Acad Sciences},
  doi={10.1073/pnas.1520752113}
}

@article{jackson2019emotion,
  title={Emotion semantics show both cultural variation and universal structure},
  author={Jackson, Joshua Conrad and Watts, Joseph and Henry, Teague R and List, Johann-Mattis and Forkel, Robert and Mucha, Peter J and Greenhill, Simon J and Gray, Russell D and Lindquist, Kristen A},
  journal={Science},
  volume={366},
  number={6472},
  pages={1517--1522},
  year={2019},
  publisher={American Association for the Advancement of Science},
  doi={10.1126/science.aaw8160}
}

@article{thompson2020cultural,
  title={Cultural influences on word meanings revealed through large-scale semantic alignment},
  author={Thompson, Bill and Roberts, Se{\'a}n G and Lupyan, Gary},
  journal={Nature Human Behaviour},
  volume={4},
  number={10},
  pages={1029--1038},
  year={2020},
  publisher={Nature Publishing Group UK London},
  doi={10.1038/s41562-020-0924-8}
}

@article{lewis2023local,
  title={Local similarity and global variability characterize the semantic space of human languages},
  author={Lewis, Molly and Cahill, Aoife and Madnani, Nitin and Evans, James},
  journal={Proceedings of the National Academy of Sciences},
  volume={120},
  number={51},
  pages={e2300986120},
  year={2023},
  publisher={National Acad Sciences},
  doi={10.1073/pnas.2300986120}
}

@article{beckner2009language,
  title = {Language is a complex adaptive system: Position paper},
  author = {Beckner, Clay and Blythe, Richard and Bybee, Joan and Christiansen, Morten H. and Croft, William and Ellis, Nick C. and Holland, John and Ke, Jinyun and Larsen-Freeman, Diane and Schoenemann, Tom},
  journal = {Language Learning},
  volume = {59},
  number = {s1},
  pages = {1--26},
  year = {2009},
  doi = {10.1111/j.1467-9922.2009.00533.x}
}

@article{pagel2009human,
  title={Human language as a culturally transmitted replicator},
  author={Pagel, Mark},
  journal={Nature Reviews Genetics},
  volume={10},
  number={6},
  pages={405--415},
  year={2009},
  publisher={Nature Publishing Group UK London},
  doi={10.1038/nrg2560}
}

@article{lenci2018distributional,
  author = {Lenci, Alessandro},
  title = {Distributional Models of Word Meaning},
  journal = {Annual Review of Linguistics},
  volume = {4},
  pages = {151--171},
  year = {2018},
  doi = {10.1146/annurev-linguistics-030514-125254},
  url = {https://www.annualreviews.org/content/journals/10.1146/annurev-linguistics-030514-125254}
}

@article{beinborn2020semantic,
  author = {Beinborn, Lisa and Choenni, Rochelle},
  title = {Semantic Drift in Multilingual Representations},
  journal = {Computational Linguistics},
  volume = {46},
  number = {3},
  pages = {571--603},
  year = {2020},
  doi = {10.1162/coli_a_00382}
}

@article{kriegeskorte2008representational,
  title={Representational similarity analysis -- connecting the branches of systems neuroscience},
  author={Kriegeskorte, Nikolaus and Mur, Marieke and Bandettini, Peter},
  journal={Frontiers in Systems Neuroscience},
  volume={2},
  pages={4},
  year={2008},
  publisher={Frontiers Media},
  doi={10.3389/neuro.06.004.2008}
}

@article{jager2018global,
  title={Global-scale phylogenetic linguistic inference from lexical resources},
  author={J{\"a}ger, Gerhard},
  journal={Scientific Data},
  volume={5},
  pages={180189},
  year={2018},
  publisher={Nature Publishing Group},
  doi={10.1038/sdata.2018.189}
}

@online{wiktionaryswadeshlists,
  author       = {{Wiktionary contributors}},
  title        = {Swadesh lists by language},
  organization = {Wiktionary},
  url          = {https://en.wiktionary.org/wiki/Category:Swadesh_lists_by_language},
  urldate      = {2026-10-22}
}

@online{wiktionaryenglishswadesh,
  author       = {{Wiktionary contributors}},
  title        = {Appendix: English Swadesh list},
  organization = {Wiktionary},
  url          = {https://en.wiktionary.org/wiki/Appendix:English_Swadesh_list},
  urldate      = {2026-10-22}
}

@misc{bowern2008conceptlist,
  author = {Bowern, Claire},
  title = {{Bowern 2008 207}},
  year = {2008},
  howpublished = {Concept list archived in Concepticon},
  url = {https://concepticon.clld.org/contributions/Bowern-2008-207}
}

@article{swadesh1955towards,
  title={Towards greater accuracy in lexicostatistic dating},
  author={Swadesh, Morris},
  journal={International Journal of American Linguistics},
  volume={21},
  number={2},
  pages={121--137},
  year={1955},
  publisher={University of Chicago Press},
  doi={10.1086/464321}
}

@inproceedings{devlin2019bert,
  title={{BERT}: Pre-training of Deep Bidirectional Transformers for Language Understanding},
  author={Devlin, Jacob and Chang, Ming-Wei and Lee, Kenton and Toutanova, Kristina},
  booktitle={Proceedings of the 2019 Conference of the North American Chapter of the Association for Computational Linguistics: Human Language Technologies, Volume 1 (Long and Short Papers)},
  pages={4171--4186},
  year={2019},
  doi={10.18653/v1/N19-1423}
}

@inproceedings{conneau2020unsupervised,
  title = {Unsupervised Cross-lingual Representation Learning at Scale},
  author = {Conneau, Alexis and Khandelwal, Kartikay and Goyal, Naman and Chaudhary, Vishrav and Wenzek, Guillaume and Guzm{\'{a}}n, Francisco and Grave, Edouard and Ott, Myle and Zettlemoyer, Luke and Stoyanov, Veselin},
  booktitle = {Proceedings of the 58th Annual Meeting of the Association for Computational Linguistics},
  year = {2020},
  pages = {8440--8451},
  doi = {10.18653/v1/2020.acl-main.747}
}

@inproceedings{vulic2020probing,
  title = {Probing Pretrained Language Models for Lexical Semantics},
  author = {Vuli{\'c}, Ivan and Ponti, Edoardo Maria and Litschko, Robert and Glava{\v{s}}, Goran and Korhonen, Anna},
  booktitle = {Proceedings of the 2020 Conference on Empirical Methods in Natural Language Processing (EMNLP)},
  pages = {7222--7240},
  year = {2020},
  doi = {10.18653/v1/2020.emnlp-main.586}
}

@article{carpenter2017stan,
  author = {Carpenter, Bob and Gelman, Andrew and Hoffman, Matthew D. and Lee, Daniel and Goodrich, Ben and Betancourt, Michael and Brubaker, Marcus and Guo, Jiqiang and Li, Peter and Riddell, Allen},
  title = {Stan: A Probabilistic Programming Language},
  journal = {Journal of Statistical Software},
  year = {2017},
  volume = {76},
  number = {1},
  pages = {1--32},
  doi = {10.18637/jss.v076.i01},
  url = {https://www.jstatsoft.org/v076/i01}
}

@article{malt1999knowing,
  title={Knowing versus naming: Similarity and the linguistic categorization of artifacts},
  author={Malt, Barbara C and Sloman, Steven A and Gennari, Silvia and Shi, Meiyi and Wang, Yuan},
  journal={Journal of Memory and Language},
  volume={40},
  number={2},
  pages={230--262},
  year={1999},
  publisher={Elsevier},
  doi={10.1006/jmla.1998.2593}
}

@book{bybee2010language,
  title={Language, usage and cognition},
  author={Bybee, Joan},
  year={2010},
  publisher={Cambridge University Press},
  doi={10.1017/CBO9780511750526}
}

@inproceedings{list2016concepticon,
  author    = {List, Johann-Mattis and Cysouw, Michael and Forkel, Robert},
  title     = {Concepticon: A Resource for the Linking of Concept Lists},
  booktitle = {Proceedings of the Tenth International Conference on Language Resources and Evaluation (LREC 2016)},
  year      = {2016},
  pages     = {2393--2400},
  url       = {https://aclanthology.org/L16-1379/}
}

@misc{haspelmath2009wold,
  author    = {Haspelmath, Martin and Tadmor, Uri},
  title     = {World Loanword Database},
  year      = {2009},
  publisher = {Max Planck Institute for Evolutionary Anthropology},
  address   = {Leipzig},
  url       = {https://wold.clld.org/}
}

@book{fellbaum1998wordnet,
  editor    = {Fellbaum, Christiane},
  title     = {WordNet: An Electronic Lexical Database},
  year      = {1998},
  publisher = {MIT Press},
  address   = {Cambridge, MA},
  doi       = {10.7551/mitpress/7287.001.0001}
}

@book{levinson2003space,
  title = {Space in Language and Cognition: Explorations in Cognitive Diversity},
  author = {Levinson, Stephen C.},
  year = {2003},
  publisher = {Cambridge University Press},
  doi = {10.1017/CBO9780511613609}
}

@book{levinson2006grammars,
  title = {Grammars of Space: Explorations in Cognitive Diversity},
  editor = {Levinson, Stephen C. and Wilkins, David P.},
  year = {2006},
  publisher = {Cambridge University Press},
  doi = {10.1017/CBO9780511486753}
}

@incollection{talmy1985lexicalization,
  title = {Lexicalization patterns: Semantic structure in lexical forms},
  author = {Talmy, Leonard},
  booktitle = {Language Typology and Syntactic Description},
  editor = {Shopen, Timothy},
  volume = {3},
  pages = {57--149},
  year = {1985},
  publisher = {Cambridge University Press},
  address = {Cambridge}
}

@article{majid2014odors,
  title = {Odors are expressible in language, as long as you speak the right language},
  author = {Majid, Asifa and Burenhult, Niclas},
  journal = {Cognition},
  volume = {130},
  number = {2},
  pages = {266--270},
  year = {2014},
  doi = {10.1016/j.cognition.2013.11.004}
}

@article{christiansen2008language,
  title = {Language as shaped by the brain},
  author = {Christiansen, Morten H. and Chater, Nick},
  journal = {Behavioral and Brain Sciences},
  volume = {31},
  number = {5},
  pages = {489--509},
  year = {2008},
  doi = {10.1017/S0140525X08004998}
}

@book{key2015ids,
  editor = {Key, Mary Ritchie and Comrie, Bernard},
  title = {The Intercontinental Dictionary Series},
  year = {2015},
  publisher = {Max Planck Institute for Evolutionary Anthropology},
  address = {Leipzig},
  url = {https://ids.clld.org/}
}

@book{croft2001radical,
  title = {Radical construction grammar: Syntactic theory in typological perspective},
  author = {Croft, William},
  year = {2001},
  publisher = {Oxford University Press},
  doi = {10.1093/acprof:oso/9780198299554.001.0001}
}

@article{church1990word,
  title = {Word association norms, mutual information, and lexicography},
  author = {Church, Kenneth W. and Hanks, Patrick},
  journal = {Computational Linguistics},
  volume = {16},
  number = {1},
  pages = {22--29},
  year = {1990}
}

@article{andrews2009integrating,
  author = {Andrews, M. and Vigliocco, G. and Vinson, D.},
  title = {{Integrating experiential and distributional data to learn semantic representations}},
  journal = {Psychological Review},
  year = {2009},
  volume = {116},
  number = {3},
  pages = {463--498},
  doi = {10.1037/a0016261}
}

@article{athanasopoulos2015two,
  author = {Athanasopoulos, P. and Bylund, E. and Montero-Melis, G. and Damjanovic, L. and Schartner, A. and Kibbe, A. and Riches, N. and Thierry, G.},
  title = {{Two languages, two minds: Flexible cognitive processing driven by language of operation}},
  journal = {Psychological Science},
  year = {2015},
  volume = {26},
  number = {4},
  pages = {518--526},
  doi = {10.1177/0956797614567509}
}

@article{fu2026semantic,
  author = {Fu, Z. and Chu, Y. and Zhang, T. and Li, Y. and Wang, X. and Bi, Y.},
  title = {{Semantic similarity across languages reflects neurocognitive dimensions shaped by climate}},
  journal = {Nature Communications},
  year = {2026},
  volume = {17},
  eid = {4016},
  doi = {10.1038/s41467-026-70608-8}
}

@article{fusaroli2012coming,
  author = {Fusaroli, R. and Bahrami, B. and Olsen, K. and Roepstorff, A. and Rees, G. and Frith, C. D. and Tylén, K.},
  title = {{Coming to terms: Quantifying the benefits of linguistic coordination}},
  journal = {Psychological Science},
  year = {2012},
  volume = {23},
  number = {8},
  pages = {931--939},
  doi = {10.1177/0956797612436816}
}

@article{kirby2008cumulative,
  author = {Kirby, S. and Cornish, H. and Smith, K.},
  title = {{Cumulative cultural evolution in the laboratory: An experimental approach to the origins of structure in human language}},
  journal = {Proceedings of the National Academy of Sciences},
  year = {2008},
  volume = {105},
  number = {31},
  pages = {10681--10686},
  doi = {10.1073/pnas.0707835105}
}

@article{landauer1997solution,
  author = {Landauer, T. K. and Dumais, S. T.},
  title = {{A solution to Plato's problem: The latent semantic analysis theory of acquisition, induction, and representation of knowledge}},
  journal = {Psychological Review},
  year = {1997},
  volume = {104},
  number = {2},
  pages = {211--240},
  doi = {10.1037/0033-295X.104.2.211}
}

@article{lewis2019distributional,
  author = {Lewis, M. and Zettersten, M. and Lupyan, G.},
  title = {{Distributional semantics as a source of visual knowledge}},
  journal = {Proceedings of the National Academy of Sciences},
  year = {2019},
  volume = {116},
  number = {39},
  pages = {19237--19238},
  doi = {10.1073/pnas.1910148116}
}

@article{liu2014language,
  author = {Liu, H.},
  title = {{Language is more a human-driven system than a semiotic system: Comment on ``Modelling language evolution: Examples and predictions'' by Tao Gong, Shuai Lan, Menghan Zhang}},
  journal = {Physics of Life Reviews},
  year = {2014},
  volume = {11},
  number = {2},
  pages = {309--310},
  doi = {10.1016/j.plrev.2013.12.008}
}

@article{liu2018language,
  author = {Liu, H.},
  title = {{Language as a human-driven complex adaptive system: Comment on ``Rethinking foundations of language from a multidisciplinary perspective'' by T. Gong et al}},
  journal = {Physics of Life Reviews},
  year = {2018},
  volume = {26--27},
  pages = {149--151},
  doi = {10.1016/j.plrev.2018.06.006}
}

@article{malikmoraleda2023concepts,
  author = {Malik-Moraleda, S. and Mahowald, K. and Conway, B. R. and Gibson, E.},
  title = {{Concepts are restructured during language contact: The birth of blue and other color concepts in Tsimane'-Spanish bilinguals}},
  journal = {Psychological Science},
  year = {2023},
  volume = {34},
  number = {12},
  pages = {1350--1362},
  doi = {10.1177/09567976231199742}
}

@article{murphy1985role,
  author = {Murphy, G. L. and Medin, D. L.},
  title = {{The role of theories in conceptual coherence}},
  journal = {Psychological Review},
  year = {1985},
  volume = {92},
  number = {3},
  pages = {289--316},
  doi = {10.1037/0033-295X.92.3.289}
}

@article{riordan2011redundancy,
  author = {Riordan, B. and Jones, M. N.},
  title = {{Redundancy in perceptual and linguistic experience: Comparing feature-based and distributional models of semantic representation}},
  journal = {Topics in Cognitive Science},
  year = {2011},
  volume = {3},
  number = {2},
  pages = {303--345},
  doi = {10.1111/j.1756-8765.2010.01111.x}
}

@article{winawer2007russian,
  author = {Winawer, J. and Witthoft, N. and Frank, M. C. and Wu, L. and Wade, A. R. and Boroditsky, L.},
  title = {{Russian blues reveal effects of language on color discrimination}},
  journal = {Proceedings of the National Academy of Sciences},
  year = {2007},
  volume = {104},
  number = {19},
  pages = {7780--7785},
  doi = {10.1073/pnas.0701644104}
}

@article{xu2025revealing,
  author = {Xu, N. and Zhang, Q. and Du, C. and Luo, Q. and Qiu, X. and Huang, X. and Zhang, M.},
  title = {{Revealing emergent human-like conceptual representations from language prediction}},
  journal = {Proceedings of the National Academy of Sciences},
  year = {2025},
  volume = {122},
  number = {44},
  eid = {e2512514122},
  doi = {10.1073/pnas.2512514122}
}

@article{zaslavsky2018efficient,
  author = {Zaslavsky, N. and Kemp, C. and Regier, T. and Tishby, N.},
  title = {{Efficient compression in color naming and its evolution}},
  journal = {Proceedings of the National Academy of Sciences},
  year = {2018},
  volume = {115},
  number = {31},
  pages = {7937--7942},
  doi = {10.1073/pnas.1800521115}
}
